%% file: aaai22.tex
\def\year{2022}\relax
\documentclass[letterpaper]{article} 
\usepackage{aaai22}  
\usepackage{times}  
\usepackage{helvet}  
\usepackage{courier}  
\usepackage[hyphens]{url}  
\usepackage{graphicx} 
\usepackage{natbib}  
\usepackage{caption} 
\DeclareCaptionStyle{ruled}{labelfont=normalfont,labelsep=colon,strut=off} 
\usepackage{algorithm}
\usepackage{algorithmic}
\usepackage{url}
\usepackage{xcolor}
\usepackage{subfigure}

\usepackage{newfloat}
\usepackage{listings}
\floatstyle{ruled}
\newfloat{listing}{tb}{lst}{}
\floatname{listing}{Listing}
\usepackage{amssymb}
\usepackage{booktabs}
\usepackage{tabularx}
\usepackage{multirow}
\usepackage{amsmath}
\usepackage{placeins}

\title{Efficient Virtual View Selection for 3D Hand Pose Estimation}

\author{
    Jian Cheng\textsuperscript{\rm 1,2}\equalcontrib, 
    Yanguang Wan\textsuperscript{\rm 1,2}\equalcontrib,
    Dexin Zuo\textsuperscript{\rm 1,2},
    Cuixia Ma\textsuperscript{\rm 1},
    Jian Gu\textsuperscript{\rm 3},\\
    Ping Tan\textsuperscript{\rm 3,4},
    Hongan Wang\textsuperscript{\rm 1},
    Xiaoming Deng\textsuperscript{\rm 1}\thanks{indicates corresponding author.},
    Yinda Zhang\textsuperscript{\rm 5\dag}
}

\affiliations{
\textsuperscript{\rm 1}Institute of Software, Chinese Academy of Sciences\\
\textsuperscript{\rm 2}University of Chinese Academy of Sciences \\ \textsuperscript{\rm 3}Alibaba \\ \textsuperscript{\rm 4}Simon Fraser University \\
\textsuperscript{\rm 5}Google \\
\{chengjian19, wanyanguang17\}@mails.ucas.ac.cn, \{xiaoming, cuixia, hongan\}@iscas.ac.cn, yindaz@google.com
}

\begin{document}

\maketitle

\begin{abstract}
3D hand pose estimation from single depth is a fundamental problem in computer vision, and has wide applications.
However, the existing methods still can not achieve satisfactory hand pose estimation results due to view variation and occlusion of human hand. In this paper, we propose a new virtual view selection and fusion module for 3D hand pose estimation from single depth.
We propose to automatically select multiple virtual viewpoints for pose estimation and fuse the results of all and find this empirically delivers accurate and robust pose estimation.
In order to select most effective virtual views for pose fusion, we evaluate the virtual views based on the confidence of virtual views using a light-weight network via network distillation.
Experiments on three main benchmark datasets including NYU, ICVL and Hands2019 demonstrate that our method outperforms the state-of-the-arts on NYU and ICVL, and achieves very competitive performance on Hands2019-Task1, and our proposed virtual view selection and fusion module is both effective for 3D hand pose estimation. 
\end{abstract}

\section{1. Introduction}
Hand pose estimation plays a key role in many applications to support human computer interaction, such as autonomous driving, AR/VR, and robotics \cite{Erolsurvey}.
Given an input image, the goal of hand pose estimation is to estimate the location of hand skeleton joints.
While many works take color images as input, methods built upon depth images usually exhibit superior performance \cite{sun2015cascaded}. 
Prior arts often use depth image and 2D CNN networks to regress 3D hand joints, or apply point-net based models \cite{qi2017pointnet,ge2018_Point} on point clouds converted from the depth using camera intrinsic parameters.
Although great progress on hand pose estimation from depth has been made in the past decade, the existing methods still can not achieve satisfactory hand pose estimation results due to the severe viewpoint variations and occlusions caused by articulated hand pose.

\begin{figure}[h]
\centering 
\includegraphics[width=\linewidth]{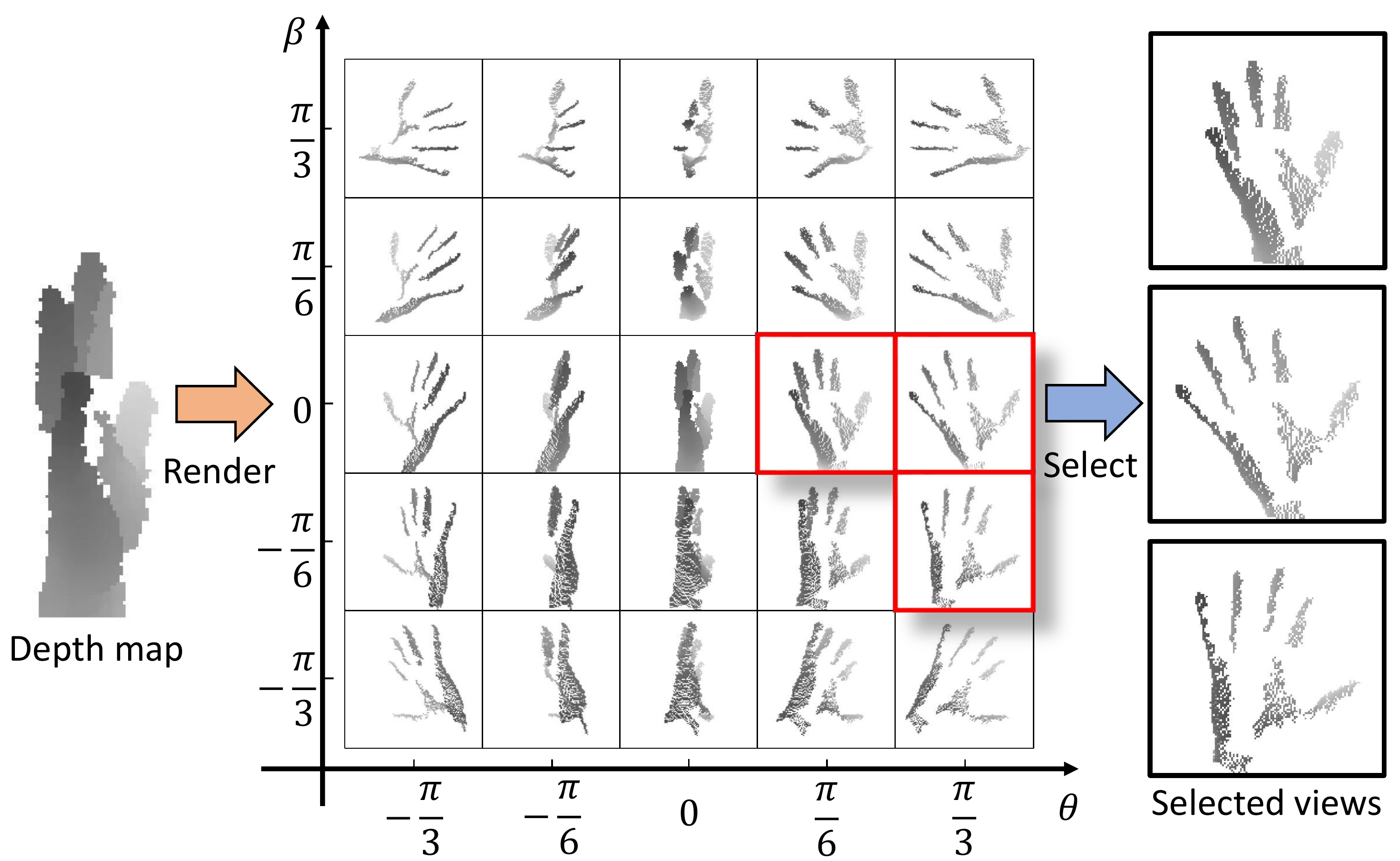}
\caption{Illustration of view selection for 3D hand pose estimation. The view of the original depth image may not be suitable for pose estimation. We select suitable views for pose estimation from the uniformly sampled virtual views. $\theta$ and $\beta$ represent the azimuth angle and the elevation angle of candidate virtual views, respectively.}
\label{fig:view_selected}
\end{figure}

\begin{figure*}[t]
\centering 
\includegraphics[width=\linewidth]{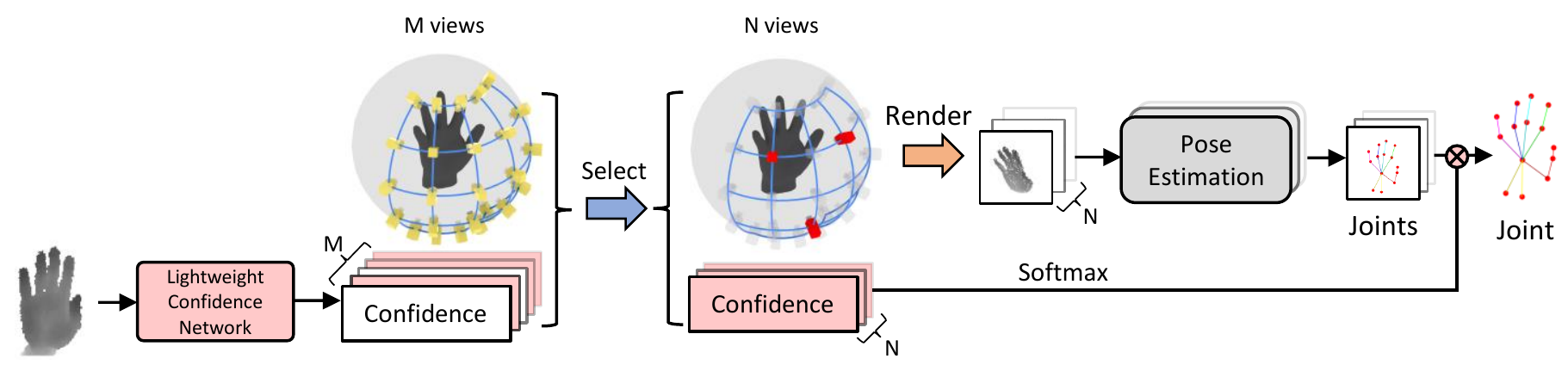}
\caption{Illustration of our virtual view selection and fusion pipeline for 3D hand pose estimation.}
\label{fig:pipeline}
\end{figure*}

To address the occlusion and viewpoint variation challenges, existing methods often rely on data alignment that transforms the input depth input into a canonical space.
However, this process is either done in 2D space \cite{sun2015cascaded,ye2016spatial} that do not fully respect the 3D nature of the depth image, or in hand-crafted canonical space, e.g. via PCA \cite{ge2018_Point} or axis-aligned bounding box \cite{ge2016robust}, that are not engaged in a joint optimization with the full model and thus the performance may not be at its best.
In contrast, learning based data alignment is more optimal as demonstrated in many previous work \cite{jaderberg2015spatial}, and back to the domain of pose estimation, this is mostly achieved in automated best viewpoint selection in multiple camera system \cite{sminchisescu2019domes,gartner2020deep} via reinforcement learning.
However, in the scenario when only one single depth is available, reinforcement learning does not typically perform well due to the limited inputs, and there is limited research study if viewpoint selection is necessary and possible with a single input depth.

We claim that viewpoint selection is very important for 3D hand pose estimation even from just a single depth.
As a perspective projection of the 3D geometry, a depth image can be projected into the 3D space as a point cloud and rendered into depth maps from another viewpoints for 3D hand pose estimation.
On one hand, the same amount of error on 2D perspective views may correspond to dramatically different errors in 3D space.
One the other hand, machine learning models may favor some typical viewpoints than the others (See Fig.~\ref{fig:view_selected} for an example).
Therefore, finding the proper virtual viewpoint to re-project the given single input depth could be critical to further improve hand pose estimation, which has been well-studied in terms of architecture.

In this paper, we propose a novel virtual view selection and fusion module for 3D hand pose estimation methods from single depth, 
which can be easily integrated into the existing hand pose estimation models to enhance the performance.
Instead of selecting just one best viewpoint, we propose to automatically select virtual viewpoints for pose estimation and fuse the results of all views, and find this empirically delivers accurate and robust pose estimation.
To achieve this, we re-render the input depth into all candidate viewpoints, and train a viewpoint selection network to evaluate the confidence of each virtual viewpoint for pose estimation.
We find this empirically works well but slows down the run-time speed when the viewpoint candidate pool is large.
To alleviate the computation issue, we adopt network distillation and show that it is possible to predict the view confidence without explicitly re-project the input depth.

The contributions of our method can be summarized as follows: We propose a novel deep learning network to predict 3D hand pose estimation from single depth, which renders the point cloud of the input depth to virtual multi-view, and get 3D hand pose by fusing the 3D pose of each view. 
We then show that the view selection can be done efficiently without sacrificing run-time via network distillation.
Extensive experiments on hand pose benchmarks demonstrate that our method achieves the state-of-the-art performance. 
The code is available in project webpage \textcolor{blue}{\url{https://github.com/iscas3dv/handpose-virtualview}}.

\section{2. Related Work}

The 3D hand pose estimation methods take depth images as input and estimate the locations of hand skeleton joints. Prior arts include \cite{moon2018v2v, ge2018point, rad2018feature, wan2018dense, xiong2019a2j}. \cite{ge2018point}, \cite{deng2020weakly} and \cite{moon2018v2v} use 3D representation of depth to estimate the hand pose, and \cite{wan2018dense}, \cite{rad2018feature} and \cite{xiong2019a2j} use 2D representation of depth to get the hand pose. 
Anchor-to-Joint Regression Network (A2J) \cite{xiong2019a2j} can 
predict accurate pose with efficiency, which sets up dense anchor points on the image and obtains the final joint locations by weighted joint voting of all anchor points. 
Although impressive results are obtained with these prior arts, these networks perform worse under occlusion or severe viewpoint variation.

The most related work to us is Ge et al. \cite{ge2016robust}, which shares the same thoughts with us that the input viewpoint may not be ideal and projects the input depth into the front, side and top view for pose estimation.
However, the number of the selected virtual views is fixed (i.e. 3), and the view selection strategy is hand-crafted but not trained in an end-to-end manner.
Different to them, we proposed a learnable virtual view selection and fusion module for 3D hand pose estimation, which can adaptively select informative virtual viewpoints 
for point cloud rendering, and fuse the estimated 3D poses of these views.

In the realm of pose estimation, viewpoint selection is also achieved in multiple camera system using reinforcement learning \cite{sminchisescu2019domes, gartner2020deep}. These works use reinforcement learning methods to select a view sequence suitable for 3D pose estimation. However, they all require a multi-view capture setup, and cannot be used for a single input depth image. Moreover, these methods are time-consuming, because views are selected in sequence, and thus reduce the inference efficiency.

\section{3. Approach}

\subsection{3.1 Overview}
In this section, we introduce our virtual view selection and fusion approach for 3D hand pose estimation from a single depth image (Fig.~\ref{fig:pipeline}). 

We first convert a depth image into 3D point clouds, and uniformly set up candidate virtual views on the spherical surface centered in the hand point clouds.
The point cloud is then rendered into candidate views as depth maps, which are then fed into a network to predict the confidence.
A 3D pose estimation network then predicts the 3D hand pose from view with top-$N$ confidence \footnote{N is a hyper-parameter that controls the run-time efficiency and pose accuracy.}, and finally fuses the pose with regard to the confidence to achieve accurate 3D hand pose results. 
To reduce the computational cost, we also design an efficient lightweight network by model distillation to predict the view confidence from the input depth itself, which saves the computation cost of point cloud rendering if the pool of candidate view is large.

\subsection{3.2 Virtual Multi-View Hand Pose Estimation}
\begin{figure}[h]
\centering 
\includegraphics[width=\linewidth]{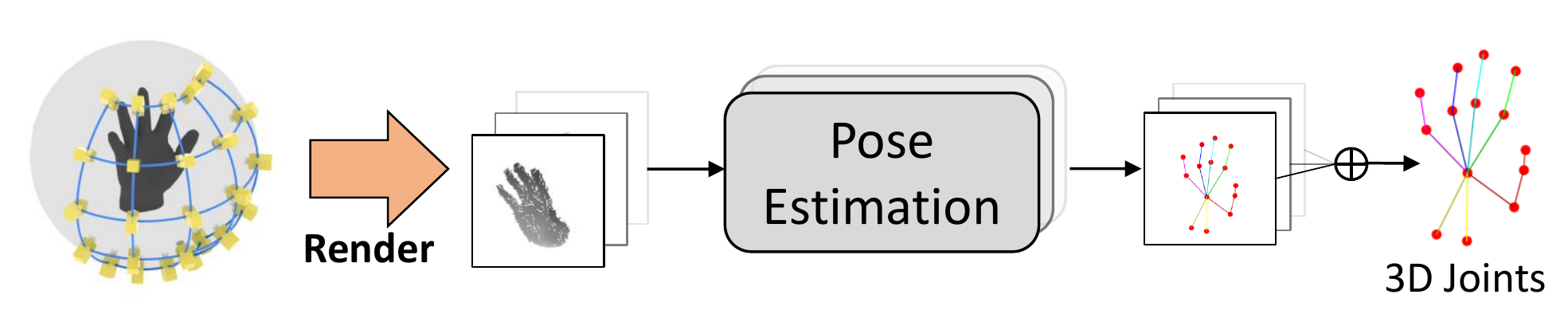}
\caption{Our virtual multi-view hand pose estimation baseline.}
\label{fig:baseline}
\end{figure}

We first explain the idea of virtual multi-view hand pose estimation.
Inspired by Ge et al. \cite{ge2016robust} that the original camera view may not be optimal for hand pose estimation, we hypothesize that denser virtual view sampling should be more beneficial and propose to exploit rendered depth on multiple virtual views to estimate 3D hand pose (Fig.~\ref{fig:baseline}).

\subsubsection{Candidate Virtual Views}
We first define a set of virtual camera views to re-render the input depth map.
Specifically, we first convert the depth map of the hand into point clouds, and uniformly sample a series of virtual cameras on the spherical surface centered in the hand point clouds.
Note that large camera rotation may cause severe  occlusion issue for depth rendering, and thus we only keep the cameras close to the input camera view.
In practice, we keep 25 virtual cameras uniformly sampled from 
the zenith angle in $[-\pi/3,\pi/3]$ and the azimuth angle in $[-\pi/3,\pi/3]$ on the sphere.  Refer to Fig.~\ref{fig:uniformly_sample} in Sec.~4.3 for illustration.

\subsubsection{Virtual View Depth Map Rendering}
The point cloud from the input depth image can now be re-projected to each of the virtual cameras.
Note that since the depth images does not provide a complete hand shape, the rendered depth may be partially incomplete or contain wrong occlusion.
However, we found empirically that these does not confuse the pose estimation network when the virtual camera is not too far from the input. Therefore, we did not complete the rendered image as the face rendering methods \cite{zhou2020rotate, fu2021high}.
We also implemented a parallel rendering process using CUDA to speed up the rendering process. 

\subsubsection{3D Hand Pose Estimation from Single Depth}
For each rendered depth image, we use the Anchor-to-Joint regression network (A2J) \cite{xiong2019a2j} as the backbone for 3D hand pose estimation for its great run-time efficiency and competitive accuracy.
In practice, any other 3D hand pose estimation models from depth can be taken as the backbone. 

We use the loss function from A2J and briefly explain here for self-contain. The loss function $L_{\textit{A2J}}$ of A2J consists of the objective loss $L_{\textit{obj}}$ and the informative anchor point surrounding loss $L_{\textit{info}}$. The objective loss $L_{\textit{obj}}$ aims to minimizing the error between the predicted hand joints with the anchor points and the ground truth hand joints, and the informative anchor point surrounding loss $L_{\textit{info}}$ aims to selecting informative anchor points located around the joints to enhance the generalization ability. The loss function $L_{\textit{A2J}}$ can be formulated as follows
\begin{equation}
\label{eq:a2j_loss}
L_{\textit{A2J}}=\lambda L_{\textit{obj}} + L_{\textit{info}}
\end{equation}
where $\lambda$ is the loss weight to balance $L_{\textit{obj}}$ and $L_{\textit{info}}$, and it is set to $\lambda=3$.

\subsubsection{Multi-View Pose Fusion}
In the end, we run a fusion stage to combine the predicted 3D hand poses from virtual camera views.
In the very basic version, 
we transform the 3D joints of each view to the original camera coordinate system with the camera extrinsic parameters, and get the final hand pose prediction by averaging the transformed hand poses.

\begin{figure}[t]
\centering 
\includegraphics[width=\linewidth]{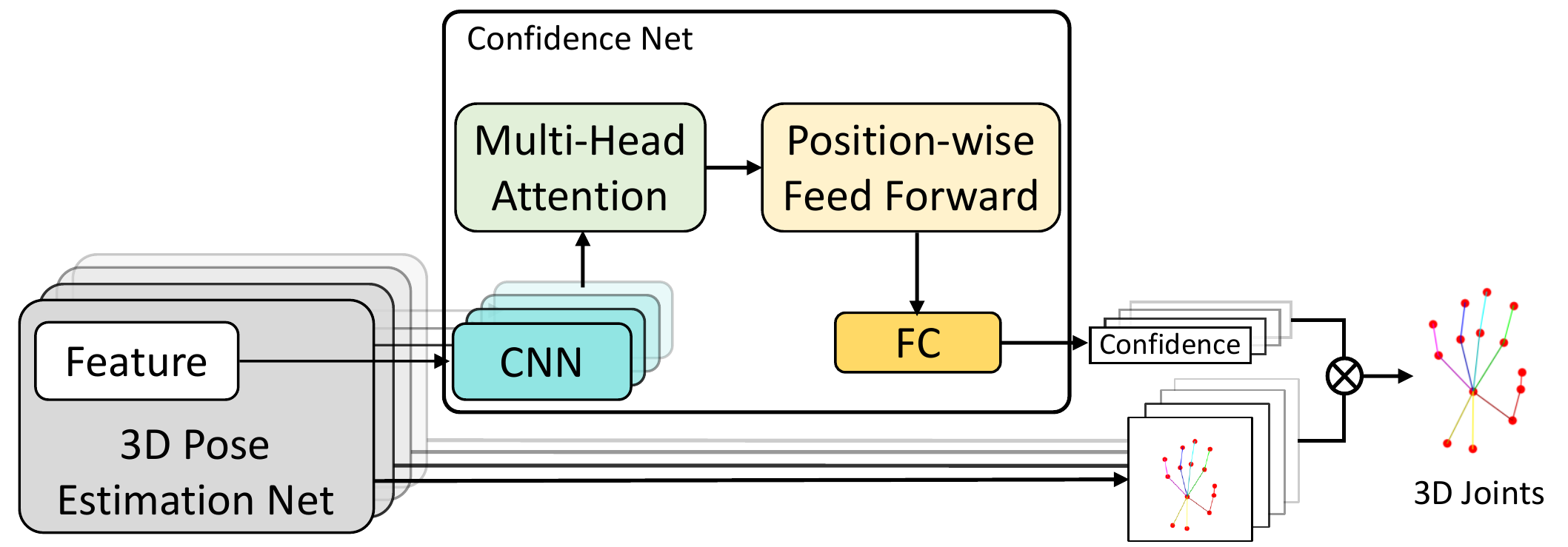}
\caption{Confidence network. The network uses CNN to extract features from the intermediate features of pose estimation net, and then uses multi-head attention to fuse the multi-views visual features. We use FC and Softmax to map the fused multi-view features to confidence of each view. 
}
\label{fig:confidence}
\end{figure}

\begin{figure*}[ht]
\centering 
\includegraphics[width=0.89\linewidth]{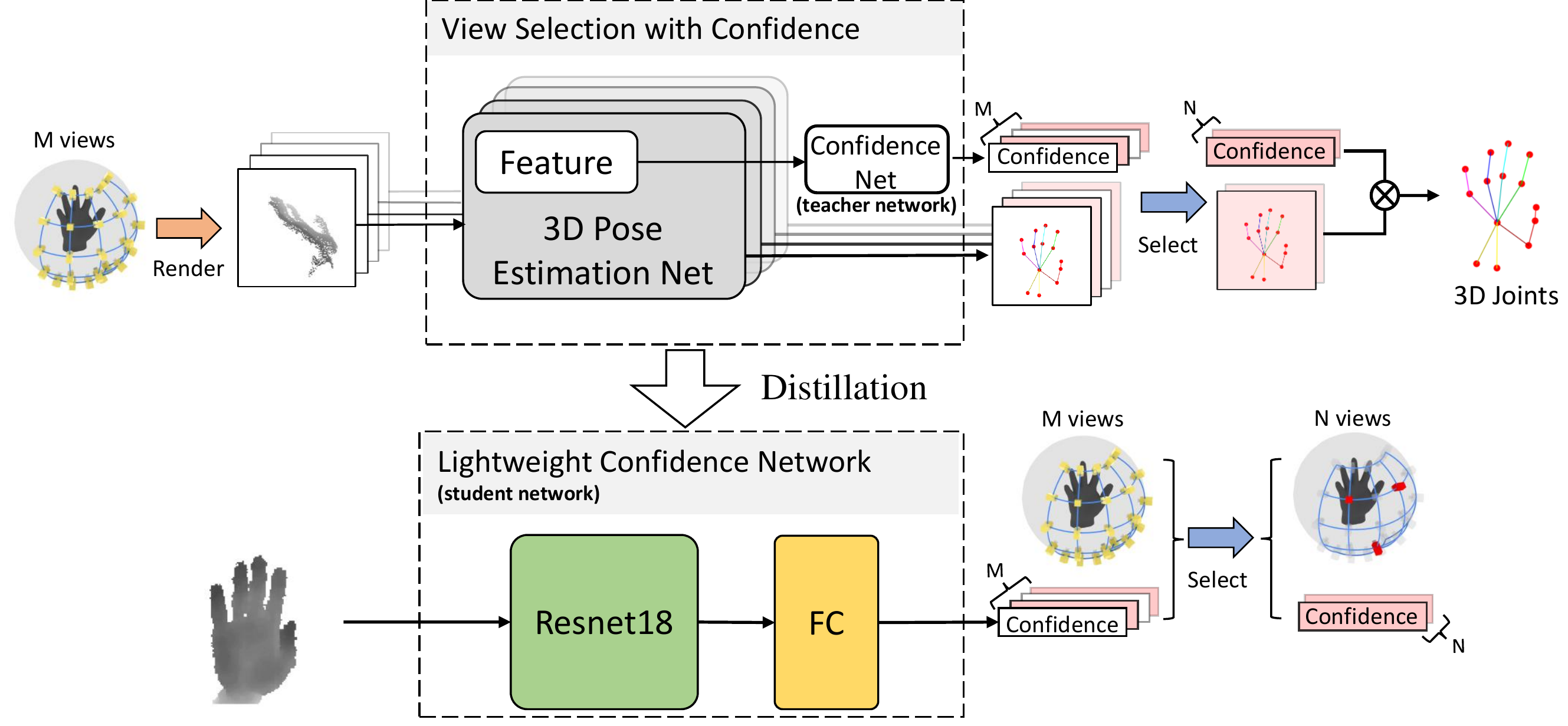}
\caption{View selection with confidence. We use multi-view depth map data to train a ``teacher'' confidence network for view selection based on confidence (top), and we train a ``student'' lightweight confidence network for efficient view selection through network distillation (bottom).}
\label{fig:multi-view_fusion}
\end{figure*}

\subsection{3.3 Virtual View Selection}
\label{sec3.3}
The method proposed in the previous section indeed significantly improves the hand pose estimation accuracy, however suffers from run-time efficiency issue.
Essentially, the virtual multi-view baseline runs single view pose estimation network multiple times, thus needs linearly proportional computation power w.r.t the size of virtual camera pool size (e.g. 25 in our case).
In another perspective, the pose prediction on some virtual views may not be great, e.g. due to sparsity or wrong occlusion, and may further hurt the overall performance when fused with an average.
Therefore, we propose a view selection algorithm that can choose a small number of high impacting camera views without the necessity of running all of them into the pose estimation network.

\subsubsection{Confidence Prediction}
The key of our view selection algorithm is a confidence network that evaluates the importance of each candidate virtual view (Fig.~\ref{fig:confidence}).
To save the computation cost, the confidence network takes the high-level feature from A2J as input.
It then extracts features using convolutional neural network, and then uses the multi-head attention in \cite{vaswani2017attention} to fuse multi-view features.
The multi-head attention mechanism can direct the network to focus on the views which play more important role, e.g. provide more accurate pose or complementary information for fusion. Finally, we use a fully connected layer to map the feature of each view to a scalar, i.e. the confidence of each view. 

\subsubsection{Virtual View Selection and Fusion}
With the predicted confidence for each virtual camera viewpoint, we can perform view selection by picking the views with top-$N$ 
confidence from all $M$ candidate views ($N<M$).
Empirically, larger $N$ tends to produce more accurate pose (See Sec.~4.3) but needs more computation.
In practice, we found that picking $N=3$ in $M=25$ virtual views can already beat most of the SoTA methods.

With the pose estimated from selected views,
we can fuse hand poses and get final output $\hat{\mathbf{J}}$ as:
\begin{equation}
\label{eq:weight_average}
\hat{\mathbf{J}} = \sum_{i=1}^{N} c_i (\mathbf{R}_i\hat{\mathbf{J}}_i+\mathbf{t}_i)
\end{equation}
where $\hat{\mathbf{J}}_i$ is the predicted hand joints of the $i$-th selected view, $c_i$ is the confidence of the $i$-th selected view after softmax, and $[\mathbf{R}_i,\mathbf{t}_i]$ is the known camera extrinsic parameters of the $i$-th selected view.

\subsubsection{Joint Training}
During the training stage, we do not give direct supervision to the confidence of each view since they are simply unknown.
Instead, we jointly train confidence network with the pose estimation network, which is supervised only on the final pose accuracy.
The joint loss $L_{J}$ is formulated as:
\begin{equation}
\label{eq:conf_loss}
L_{J}=\sum_{i=1}^K L_{\tau}( \Vert \mathbf{J}_i-\hat{\mathbf{J}}_i \Vert)
\end{equation}
where $\hat{\mathbf{J}}_i$ is the estimated $i$-th joint by multi-view pose fusion (See Eq.~(\ref{eq:weight_average})), $\mathbf{J}_i$ is the ground truth location of the $i$-th joint, $K$ is the number of hand joint, and $L_{\tau}(\cdot)$ is the smooth$_{L1}$ loss function. 

Note that the behavior of confidence network might be different with varying $N$ since the predicted confidences have been used at both selection and fusion stage.
For the selection stage, only the ranking of the confidence matters; and for the fusion stage, the precise value of confidences of the chosen views also matters.
Therefore, for ideal performance, the confidence network should be trained for each specific $N$.
In practice, however, a single model trained with $N=M$ can still work for most of the varying $N$, with slightly worse but still reasonable performance.

\subsubsection{Distillation for Efficient Confidence Prediction}
With view selection, we are able to reduce the number of forward pass of the pose estimation network from the total number of virtual views $M$ to a much smaller number of selected views $N$.
This reduces enormous computation, but we still observe noticeable frame per second (FPS) drops mostly due to two reasons: 1) The input depth still needs to be rendered to all candidate camera views for confidence estimation. 2) The confidence network still needs to process all the rendered depth, though designed to be light-weighted but still costly to run the multi-head attention.

We resort to network distillation to alleviate this run-time issue.
Specifically, we take our confidence network as the teacher network, and train an even more light-weighted student network (a ResNet-18 followed by a fully connected layer).
More importantly, the student network takes only the original depth as the input and directly outputs confidence for all $M=25$ views.
This effectively removes the necessity of re-rendering the input depth to all virtual cameras.

To train the student network, we directly take the confidence predicted by teacher network as the ground truth.
The loss is defined as 
\begin{equation}
\label{eq:light_loss}
L_{\textit{light}} = \sum_{i = 1}^M L_{\tau}(\beta (c_{i} - {\hat{c}}_{i} ) )
\end{equation}
where $\beta$ is the scaling factor set to $100$, ${\hat{c}}_{i} $ is the confidence ground truth from teacher network, i.e. the multi-head attention network from $M$ views, and $c_{i}$ is the student network output.

Once the training is done, the teacher network is no longer needed during the inference, and student network can support the view selection and fusion in an efficient way.

\subsection{3.4 Implementation Details}
We train and evaluate our models on a workstation with two Intel Xeon Silver 4210R, 512GB of RAM and an Nvidia RTX3090 GPU. Our models are implemented within PyTorch. Adam optimizer is used; the initial learning rate is set to 0.001 and is decayed by 0.9 per epoch.  In order to conduct data augmentation, we randomly scale the cropped depth map, jitter the centroid of the point cloud, and randomly rotate the camera when rendering multi-view depth. For all smooth$_{L1}$ loss, the switch point between quadratic and linear is set to 1.0.

Our network consists of the 3D pose estimation network (i.e. A2J), the teacher confidence network and the lightweight student confidence network. 
The network input is $176 \times 176$ hand region cropped from the input depth, and we use ResNet-50 as the backbone of A2J.
We first train the 3D pose estimation network and the teacher confidence network together, the loss can be formulated as:
\begin{equation}
\label{eq:final_loss}
L_{\textit{viewsel}} = L_{\textit{A2J}} + \gamma L_J
\end{equation}
where $\gamma = 0.1$ is the factor to balance the loss terms.

Then we fix the parameters of the two networks and train the lightweight student confidence network with $L_{light}$.

\section{4. Experiments}
\subsection{4.1 Datasets and Evaluation Metric}
\paragraph{NYU Hand Pose Dataset (NYU)} \cite{tompson2014real} contains 72,757 frames for training and 8,252 frames for testing. 36 hand joints are annotated,  but we use only a subset of 14 hand joints for evaluations following the same evaluation protocol in \cite{tompson2014real}.

\paragraph{ICVL Hand Pose Dataset (ICVL)} \cite{tang2014latent} contains 331,006 frames for training and 1,596 frames for testing. 16 hand joints are annotated.

\paragraph{Task 1 of Hands19 Challenge Dataset (Hands19-Task1)} \cite{armagan2020measuring} contains 175,951 training depth images from 5 subjects and 124,999 testing depth images from 10  subjects, in which 5 subjects overlap with the training set. This dataset is very challenging because of its exhaustive coverage of viewpoints and articulations.

\paragraph{Evaluation Metric} We evaluate the hand pose estimation performance using standard metrics proposed in \cite{tang2014latent}, i.e. mean joint error and the percentage of test examples that have all predicted joints within a given maximum distance from the ground truth.

\subsection{4.2 Comparison with State-of-the-art Methods}

\begin{table}[h]
	\centering
	\resizebox{\linewidth}{!}{
	\begin{tabular}{ccc}
		\toprule[1pt]
		Methods & NYU & ICVL \\ 
		\midrule
		HandPointNet \cite{ge2018_Point} & 10.54 & 6.93 \\
		DenseReg \cite{wan2018dense} & 10.21 & 7.24 \\
		P2P \cite{ge2018point} & 9.05 & 6.33 \\
		A2J \cite{xiong2019a2j} & 8.61 & 6.46 \\
		V2V \cite{moon2018v2v} & 8.42 & 6.28 \\
		AWR \cite{huang2020awr} & 7.37 & 5.98 \\
		\hline
		Ours-1view & 7.34 & 5.16 \\
		Ours-3views & 6.82 & 4.86 \\
		Ours-9views & 6.53 & 4.77 \\
		Ours-15views & 6.41 & \textbf{4.76}\\
		Ours-25views & \textbf{6.40} & 4.79\\
		\bottomrule[1pt]
	\end{tabular}
	}
	\caption{Comparison mean joint 3D error (mm) and ranking result with state-of-art methods on NYU  dataset and ICVL dataset. ``Ours-1view'', ``Ours-3views'', ``Ours-9views'' and ``Ours-15views''  are the results of our method with selected 1, 3, 9 and 15 views from 25 uniformly sampled views, respectively.
    ``Outs-25views'' denotes the results of our method with 25 uniformly sampled views.}
	\label{NYU}
\end{table}

\begin{figure}[ht]
\centering 
\includegraphics[width=\linewidth]{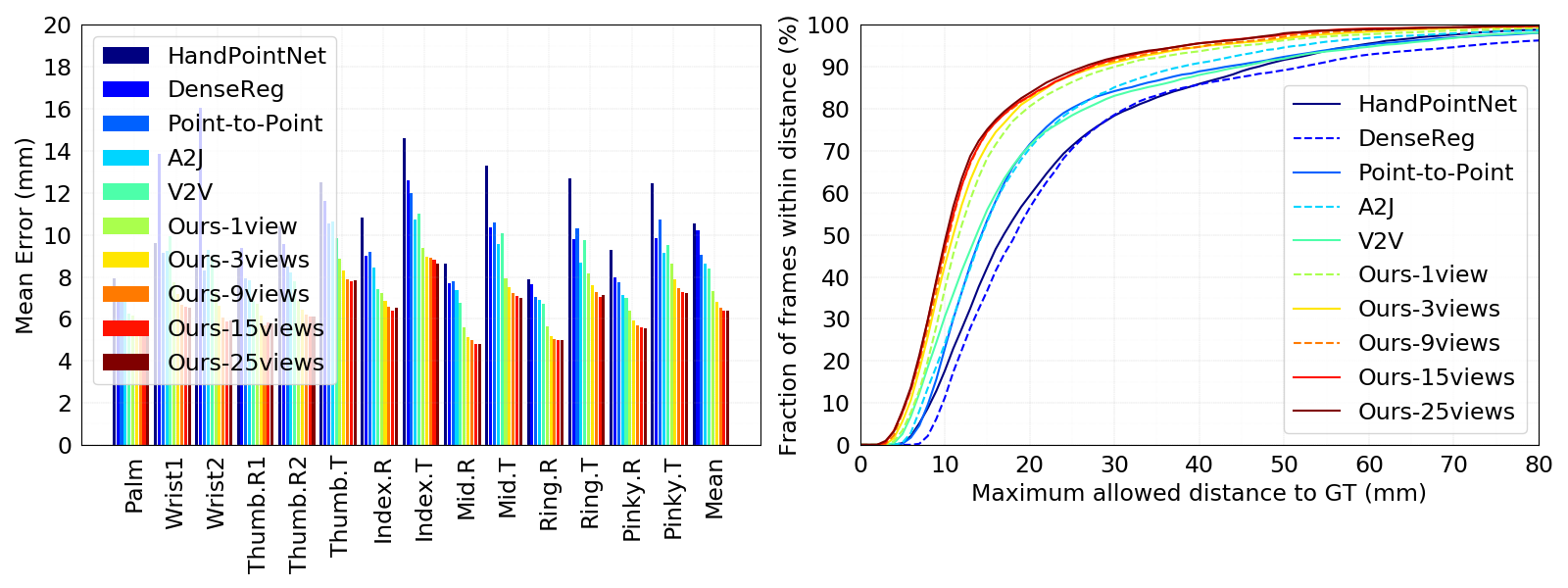}
\caption{Comparison of our proposed method with state-of-the-art methods on NYU dataset. Left: mean joint error per hand joint. Right: the percentage of success frames over different error thresholds.}
\label{fig:nyu_error}
\end{figure}

\begin{table}[h]
	\centering 
	\resizebox{\linewidth}{!}{
	\begin{tabular}{cc}  
		\toprule[1pt]
		Methods & Mean error (mm) \\  
		\midrule
		{V2V} \cite{moon2018v2v} & 15.57 \\
		{AWR} \cite{huang2020awr} & 13.76 \\
		{A2J} \cite{xiong2019a2j} & 13.74 \\ 
		{Rokid} \cite{zhang2020handaugment} & 13.66 \\
		\hline
		Ours-1view & 14.14 \\
		Ours-3views & 13.24 \\
		Ours-9views & 12.67 \\
		Ours-15views & \textbf{12.51} \\
		Ours-25views & 12.55 \\
		\bottomrule[1pt]
	\end{tabular}
	}
	\caption{Comparison with state-of-art methods on Hands19-Task1. We show the mean joint error on the test dataset split “Extrapolation”, which aims to evaluating the model generalization performance and is the main evaluation metric on Hands19-Task1. ``Ours-1view'', ``Ours-3views'', ``Ours-9views'', ``Ours-15views'' and ``Ours-25views'' have the same meaning as shown in Table~\ref{NYU}}
    \label{HANDS2019} 
\end{table}

\begin{figure*}[ht]
    \centering
    \resizebox{\linewidth}{!}{  
    \begin{tabular}{ccccccccc}
      \includegraphics[width=0.09\linewidth]{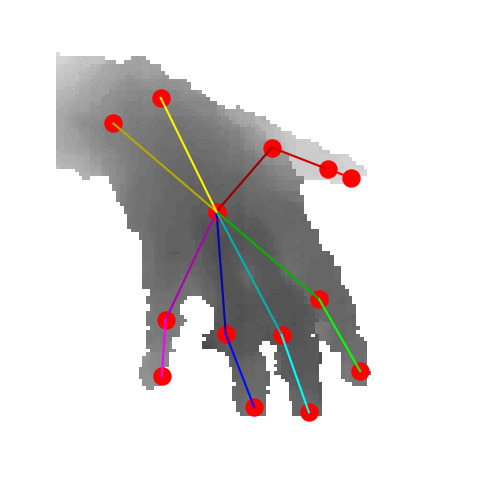}   &  \includegraphics[width=0.09\linewidth]{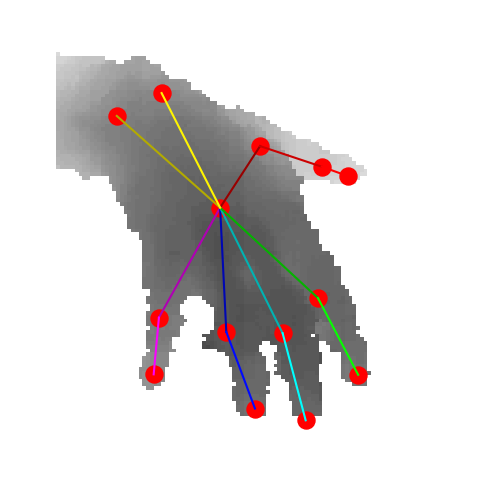} & \includegraphics[width=0.09\linewidth]{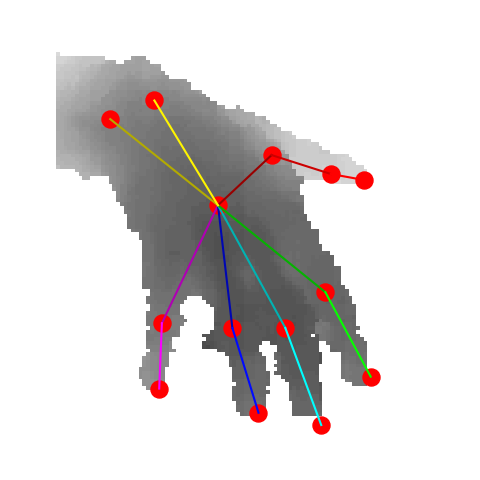} & 
      \includegraphics[width=0.09\linewidth]{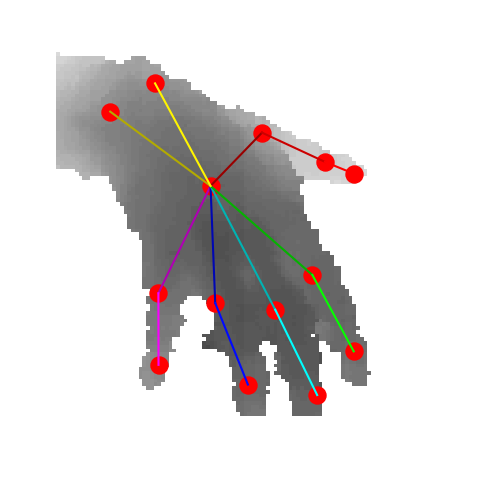}& 
      \includegraphics[width=0.09\linewidth]{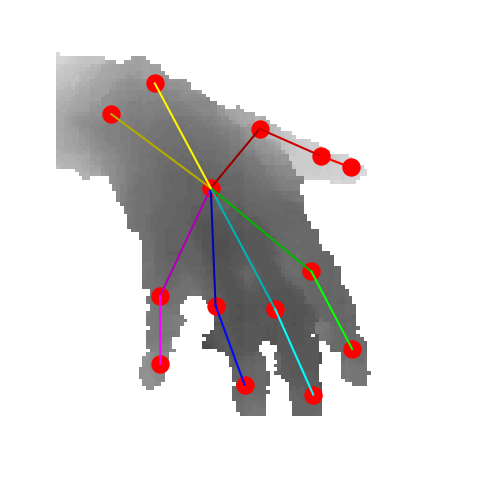}& 
      \includegraphics[width=0.09\linewidth]{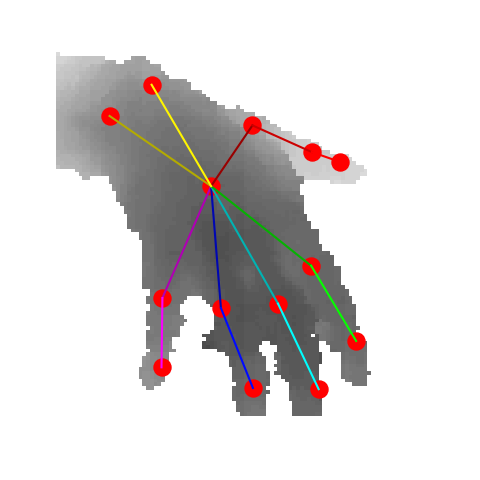}& 
      \includegraphics[width=0.09\linewidth]{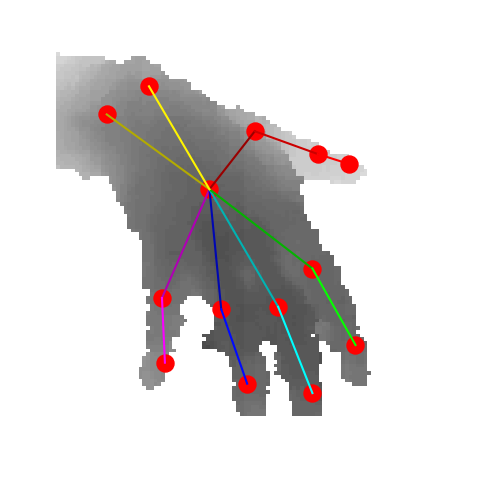}& 
      \includegraphics[width=0.09\linewidth]{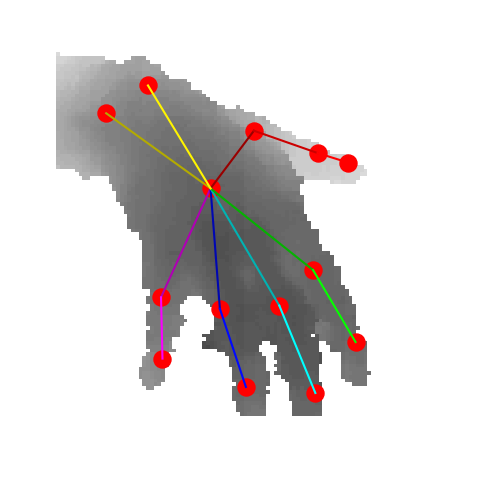}&
      \includegraphics[width=0.09\linewidth]{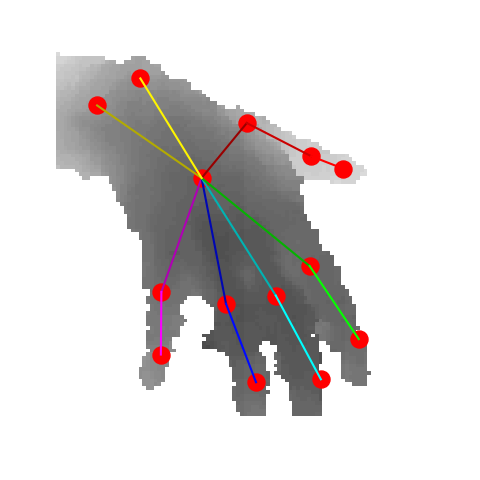} \\
      \includegraphics[width=0.09\linewidth]{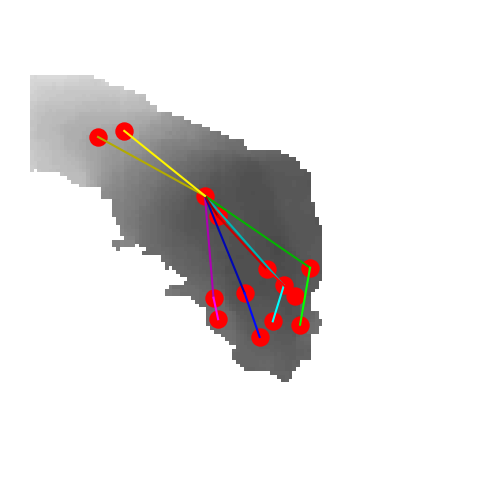}   & \includegraphics[width=0.09\linewidth]{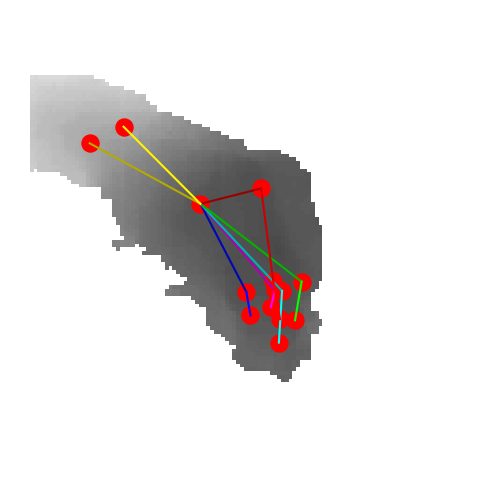}   & \includegraphics[width=0.09\linewidth]{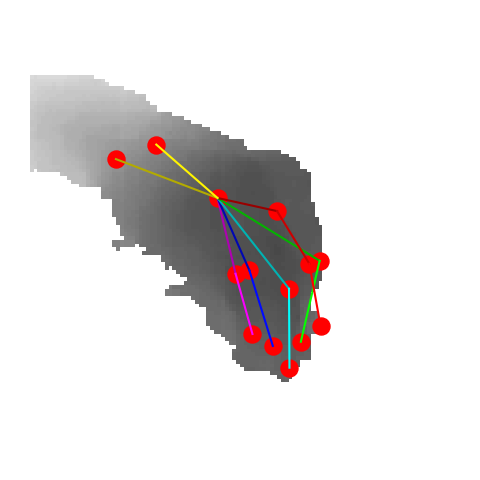}  &  \includegraphics[width=0.09\linewidth]{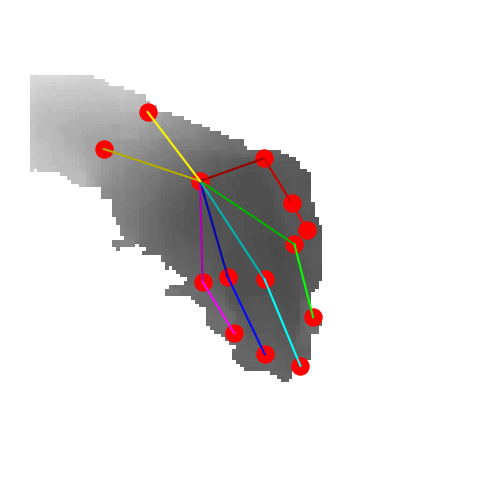}& 
      \includegraphics[width=0.09\linewidth]{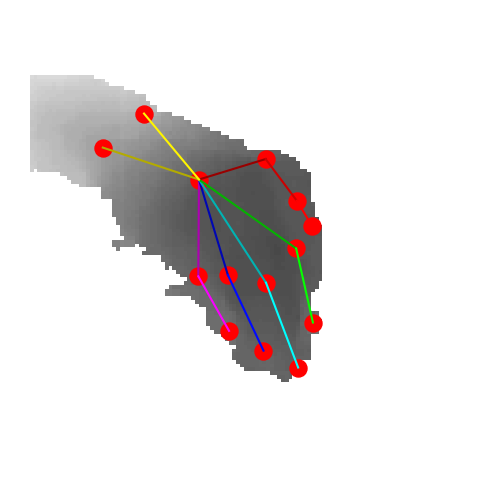}& 
      \includegraphics[width=0.09\linewidth]{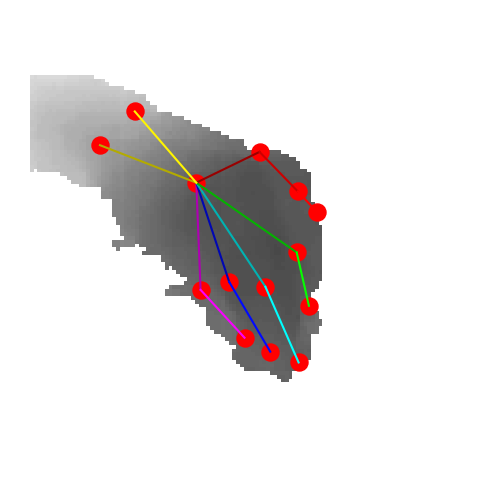}& 
      \includegraphics[width=0.09\linewidth]{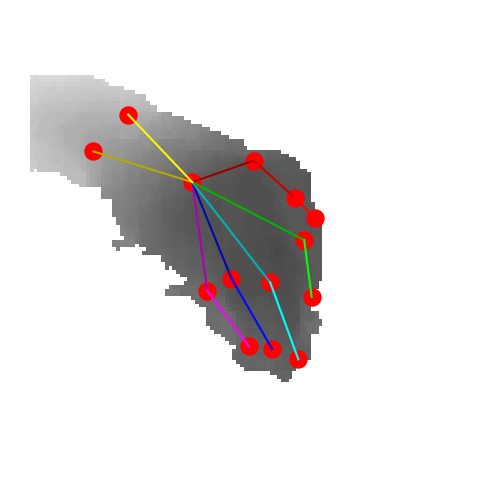}& 
      \includegraphics[width=0.09\linewidth]{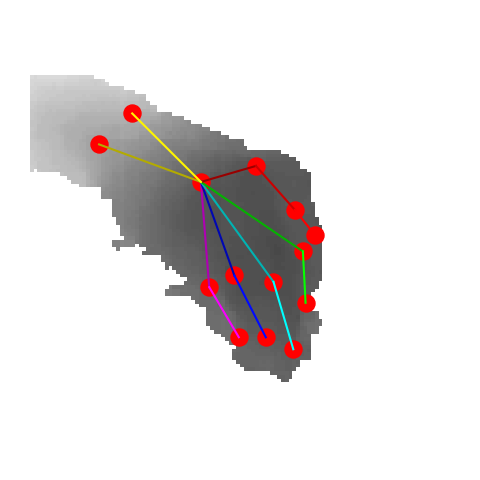}&
      \includegraphics[width=0.09\linewidth]{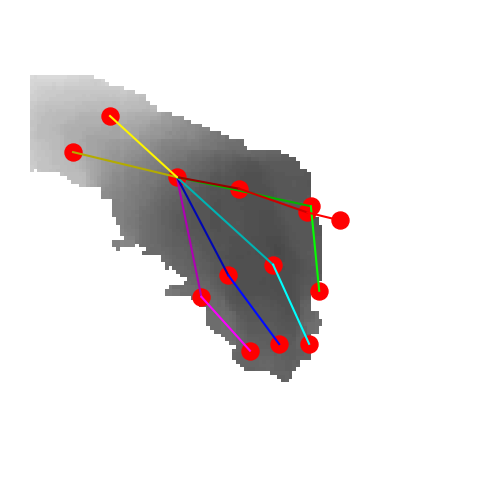} \\ 
      \includegraphics[width=0.09\linewidth]{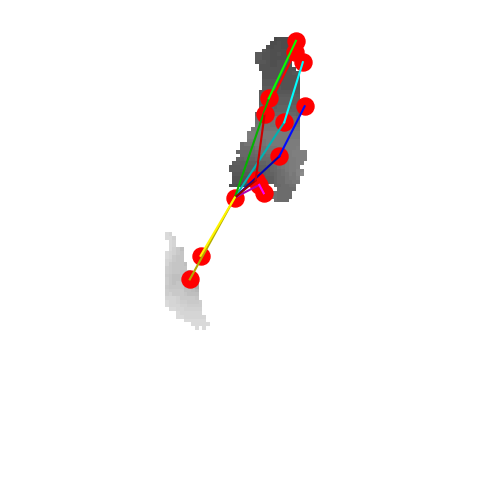} &\includegraphics[width=0.09\linewidth]{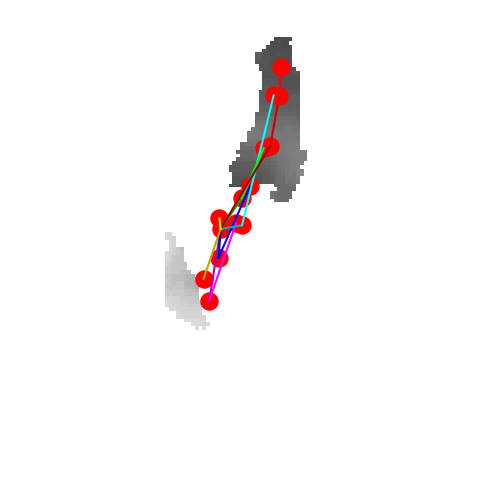} 
      &\includegraphics[width=0.09\linewidth]{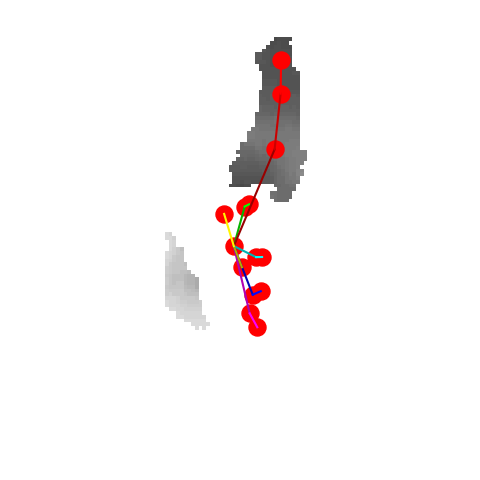} & \includegraphics[width=0.09\linewidth]{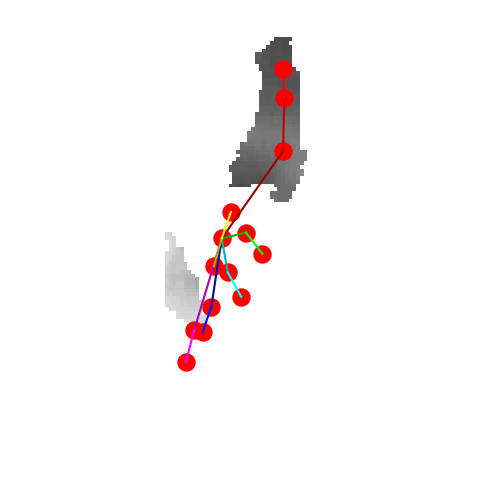}& 
      \includegraphics[width=0.09\linewidth]{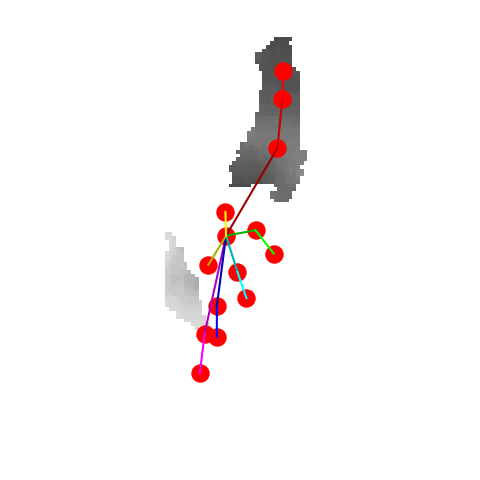}& 
      \includegraphics[width=0.09\linewidth]{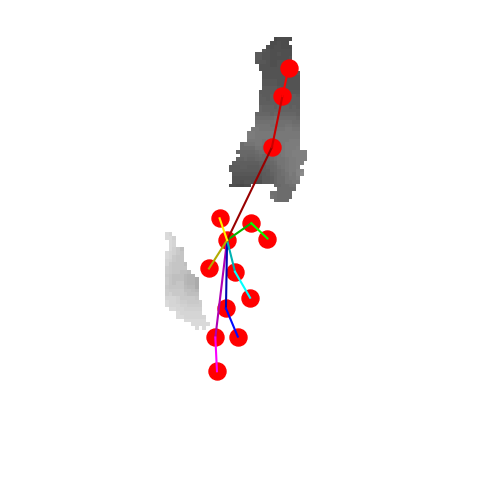}& 
      \includegraphics[width=0.09\linewidth]{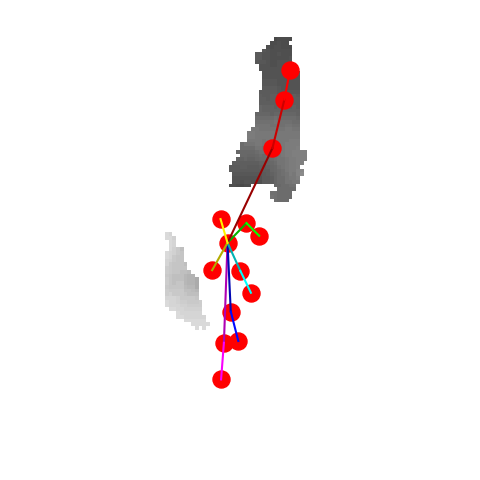}& 
      \includegraphics[width=0.09\linewidth]{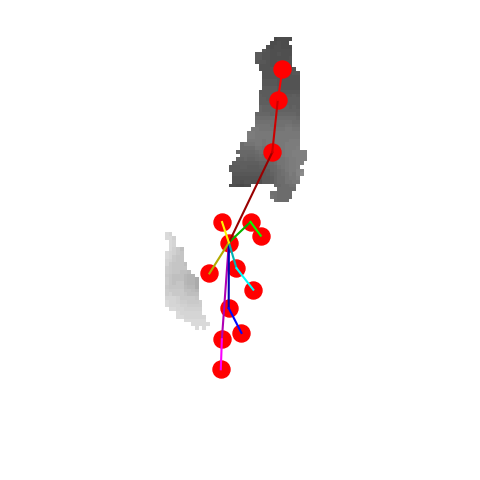}&
      \includegraphics[width=0.09\linewidth]{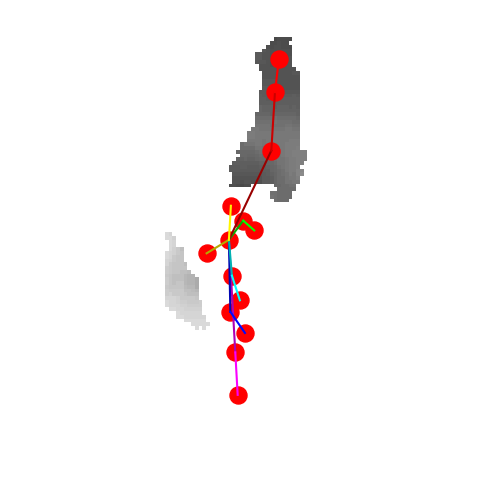} \\ 
      P2P & V2V & A2J & 1 view & 3 views & 9 views& 15 views & 25 views & Ground truth
    \end{tabular}
    }
    \caption{Comparison visualization results with state-of-art methods on NYU dataset. 
    ``1 view'', ``3 views'', ``9 views'' and ``15 views''  are the results of our method with selected 1, 3, 9 and 15 views from 25 uniformly sampled views, respectively.
    ``25 views'' denotes the results of our method with 25 uniformly sampled views.
    }
    \label{fig:comp_STOA}
\end{figure*}
In this experiment, we use our ``student'' lightweight confidence network for view selection and pose fusion.

We first compare our method with the state-of-the-art methods
on NYU dataset and ICVL dataset. 
DenseReg \cite{wan2018dense}, A2J \cite{xiong2019a2j} and AWR \cite{huang2020awr} directly use the depth map for pose estimation. HandPointNet \cite{ge2018_Point}, P2P \cite{ge2018point} and V2V \cite{moon2018v2v}\cite{hruz2021hand} use the 3D representation of the depth map for pose estimation.

Table~\ref{NYU} shows the mean joint error. Fig.~\ref{fig:nyu_error} shows the percentage of success frames over different error thresholds and the error of each joint. We do not show AWR in Fig.~\ref{fig:nyu_error} because the best prediction results on NYU dataset are not released by AWR.
Note that with just 1 view selected, we already outperform all the other methods. And when more view selected, the performance keeps improving.
This clearly indicates that our view selection is effective in finding a better virtual view for pose estimation, and more views are benefical through the confidence based fusion.

Fig.~\ref{fig:comp_STOA} shows the qualitative comparison to other methods on a few testing examples of NYU dataset.
Our method performs especially better on views with heavy occlusion and missing depth, e.g. the 3rd row.
It is as expected since the re-render in a perpendicular virtual camera looking at the palm might be better to interpret the input depth.

We also compare our method with the state-of-the-art 3D hand pose estimation methods 
on Hands19-Task1. The details and results of state-of-the-art methods are cited from \cite{armagan2020measuring}. {Rokid} \cite{zhang2020handaugment} trains 2D CNN to regress joints with additional synthetic data.
Compared to the used A2J backbone in our method, 
{A2J} \cite{xiong2019a2j} reports results using higher resolution depth as input ($384 \times 384$), deeper backbone (ResNet-152), and 10 backbone model ensemble. {AWR} \cite{huang2020awr} also provides results with model ensemble.
Table~\ref{HANDS2019} shows the performance comparison. we can observe that our method with 3 selected views outperform other methods.
Our method with 1 selected view performs slightly worse than {AWR}, {A2J} and {Rokid}, which can be due to
using model ensemble in {AWR} and {A2J} and using additional synthetic training data in {Rokid} \cite{zhang2020handaugment}.  

More experimental results can be found in the supplementary document.

\subsection{4.3 Ablation Study}

\begin{figure}[ht]
\centering 
\includegraphics[width=\linewidth]{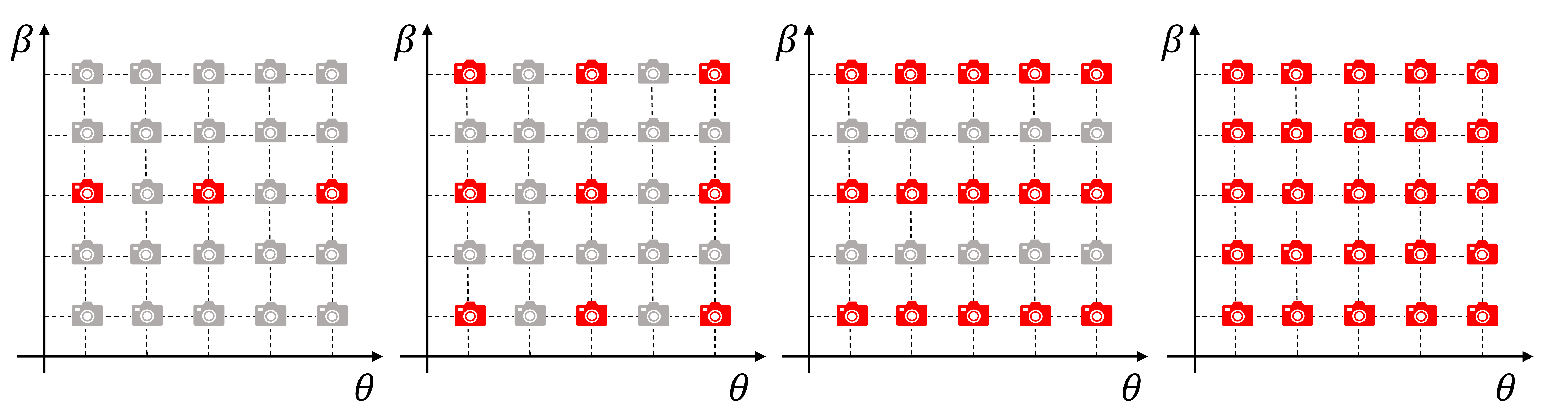}
\caption{Uniform sampling schemes. 25 virtual cameras are set uniformly. Red cameras are candidate virtual views. The figures from left to right show how 3, 9, 15, and 25 candidate virtual views are uniformly sampled.}
\label{fig:uniformly_sample}
\end{figure}

\begin{figure}[ht]
    \centering 
    \includegraphics[width=\linewidth]{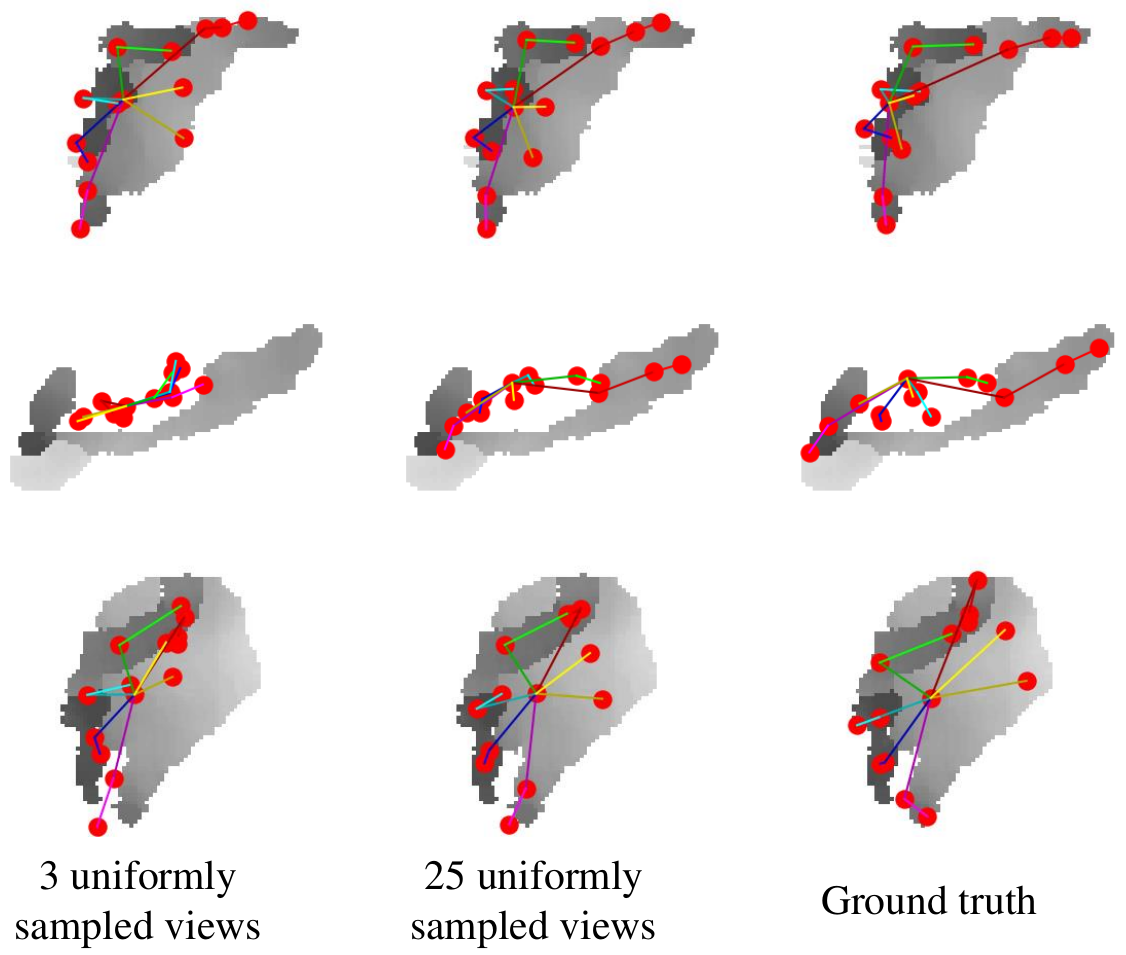}
    \caption{Comparison the visualization results of our method using different numbers of views on NYU dataset.}
    \label{fig:comp_views}
\end{figure}

\paragraph{Effect of View Number}
In order to investigate the effect of the number of virtual views with our method, we compare the hand pose performance on NYU dataset, ICVL dataset and Hands19-Task1 dataset by generating 3, 9, 15, 25 uniformly sampled virtual views (``UNIFOMR" column of Table~\ref{error_view_select}).
We also visualize the results of our method using different numbers of views on NYU dataset in Fig.~\ref{fig:comp_views}
We observe that the hand pose estimation error decreases as the number of virtual views increases. Therefore, the virtual multi-view can boost the hand pose performance. The visualization results show that when using 25 views, the estimated wrist joints and joints in the missing depth value area are more accurate than using 3 views.

\begin{table*}[ht]
	\centering
	\resizebox{\linewidth}{!}{    
	\begin{tabular}{c|ccc|ccc|ccc}
        \toprule[1pt]
        ~ & \multicolumn{3}{c|}{NYU} & \multicolumn{3}{c|}{ICVL} & \multicolumn{3}{c}{Hands2019-Task1} \\
        \midrule
        Number of views & UNIFORM & SELECT & LIGHT & UNIFORM & SELECT & LIGHT & UNIFORM & SELECT & LIGHT \\
        \midrule
        1 view & 7.93 & 7.23 & 7.34 & 5.56 & 5.18 & 5.16 & 14.39 & 14.03 & 14.14 \\
        3 views & 7.14 & 6.73 & 6.82 & 5.27 & 4.85 & 4.86 & 13.67 & 13.07 & 13.24 \\
        9 views & 6.77 & 6.43 & 6.53 & 4.96 & 4.77 & 4.77 & 12.81 & 12.60 & 12.67\\
        15 views & 6.55 & 6.38 & 6.41 & 4.85 & 4.76 & 4.76 & 12.61 & 12.51 & 12.51 \\
        25 views & 6.40 & - & - & 4.79 & - & - & 12.55 & - & - \\
        \bottomrule[1pt]
	\end{tabular}
	}
	\caption{Comparison of mean joint error using uniform sampling and view selection on NYU, ICVL and Hands2019-Task1. ``UNIFORM" denotes using uniformly sampled views. ``SELECT" denotes selecting views from 25 uniformly sampled views with the ``teacher'' confidence network. ``LIGHT" denotes selecting views from 25 uniformly sampled views with the ``student'' lightweight confidence network.}
	\label{error_view_select} 
\end{table*}

\paragraph{Effect of View Selection}
In this section, we investigate if the confidences are effective for view selection.
We use our method to select 3, 9, 15 views from 25 candidate views and compare to uniform sampling as illustrated in Fig.~\ref{fig:uniformly_sample}.
Though simple, the uniform sampling is actually a very strong strategy since it roughly guarantee at least 1 views close to some good candidate.
The results on NYU dataset, ICVL dataset and Hands19-Task1 dataset are shown in Table~\ref{error_view_select}.
For each experiment, we show the performance using teacher and student confidence network (``SELECT'': teacher network, ``LIGHT'': student network).
Our view selection consistently outperforms uniform selection in all the settings, which indicates that the predicted confidence is effective for view selection.
Though using student network results in slightly worse performance than the teacher network, the overall computation cost is significantly reduced.
As shown in Table~\ref{fps_view_select}, using student network almost doubles FPS.
On ICVL dataset, the student network performs better than the teacher network, which may be due to low annotation quality and small scale of ICVL.
More comparisons to random sampled views can be found in the supplementary document.

\begin{table}[t]
	\centering
	\resizebox{\linewidth}{!}{
	\begin{tabular}{cccc}
        \toprule[1pt]
        Number of views & UNIFORM & SELECT & LIGHT \\
        \midrule
        1 view & 61.57 & 27.92 & 47.43 \\
        3 views & 56.71 & 28.19 & 46.42 \\
        9 views & 41.56 & 27.61 & 39.48 \\
        15 views & 36.50 & 27.95 & 34.57 \\
        25 views & 26.71 & - & - \\
        \bottomrule[1pt]
	\end{tabular}
	}
	\caption{FPS comparison of uniform sampling and view selection on NYU dataset. ``UNIFORM", ``SELECT", ``LIGHT" have the same meaning as shown in Table~\ref{error_view_select}.}
	\label{fps_view_select} 
\end{table}

\paragraph{Comparison of Different Multi-View Fusion}
We now evaluate the confidence based fusion.
In Table~\ref{confidence}, we compare to direct average without the confidence on NYU dataset, ICVL dataset and Hands2019-Task1 dataset.
The confidence based fusion achieves better performance on 3 datasets than direct average, which shows that the confidence is also beneficial to guide the fusion stage.  

\begin{table}[t]
	\centering
	\resizebox{\linewidth}{!}{
	\begin{tabular}{cccc}
	    \toprule[1pt]
	    Component & NYU & ICVL & Hands2019-Task1 \\
		\midrule
	    w confidence & 6.40 & 4.79 & 12.55 \\
	    w/o confidence & 6.58 & 4.88 & 12.75 \\
		\bottomrule[1pt]
	\end{tabular}}
	\caption{Comparison of mean 3D joint error using different multi-view fusions on 25 uniformly sampled views.}
	\label{confidence} 
\end{table}

\section{5. Conclusion}
In this paper, we propose a deep learning network to learn 3D hand pose estimation from single depth, which renders the point cloud to virtual multi-view, and get more accuracy 3D hand pose by fusing the 3D pose of each view. Meanwhile, the confidence network we built uses multi-head attention to calculate the confidence of each virtual view. The confidence is used to improve the accuracy of 3D reconstruction during fusion and view selection. In order to obtain the confidence of each view efficiently, we obtain a lightweight confidence network using network distillation. Experiments on the main benchmark datasets demonstrate the effectiveness of our proposed method. 
Our method can also inspire several related researches such as scene parsing and reconstruction from depth, etc.

\vspace{5mm}
\noindent \textbf{Acknowledgments} This work was supported by National Natural Science Foundation of China (No.~61473276), Beijing Natural Science Foundation (L182052), and Distinguished Young Researcher Program, ISCAS.

\bibliography{aaai22}
\newpage
\section*{Appendix}
\input{supp}

\end{document}

%% file: supp.tex
\section{1. Network Details of Confidence Network}
In this section, we elaborate on the network architecture of the ``teacher'' confidence network in Section~3.3 and Fig.~4 of our main submission.

In order to obtain the confidence of each view, our method first uses a convolutional neural network to extract features, then uses multi-head attention \cite{vaswani2017attention} to fuse multi-view features, and finally converts the fused multi-view visual features into the confidence of each view using a fully connected layer. 

\begin{figure}[hb]
\centering 
\includegraphics[width=\linewidth]{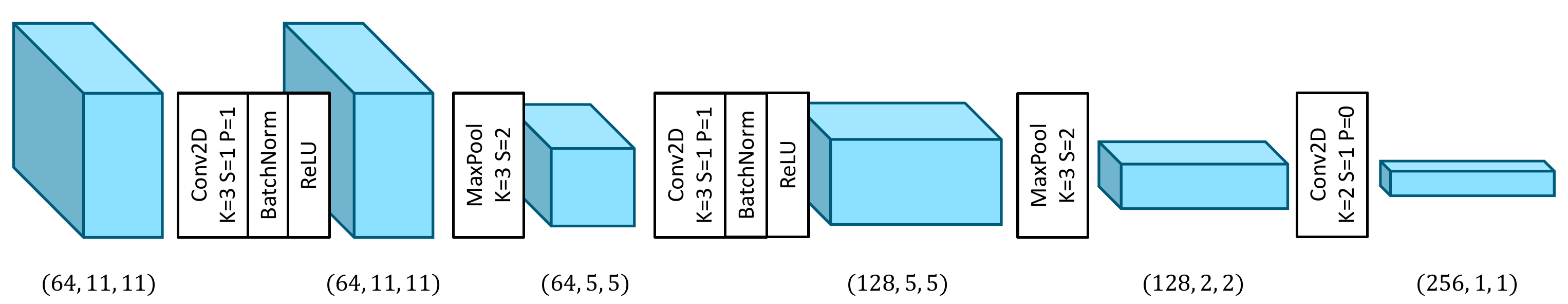}
\caption{Network details of the convolutional neural network for feature extraction. `K' stands for `kernel size', `S' stands for `stride', and `P' stands for `padding'.}
\label{fig:cnn}
\end{figure}

Fig.~\ref{fig:cnn} shows the details of network architecture of convolutional neural network for feature extraction. The network takes the low-level feature from A2J as input, and adopts 3 convolutional layers for feature encoder.
The network extracts a 256-dimensional feature for each view, and then feeds the extracted features of multiple views to a multi-head attention module \cite{vaswani2017attention} for feature fusion.
In the multi-head attention module, we 
set the number of heads $n_{\textit{head}}$ to 1, $d_q, d_k, d_v$ are the dimensions of $Q$, $K$ and $V$ (defined as \cite{vaswani2017attention}), and they are all set to 64. 

\begin{table*}[ht]
	\centering
	\resizebox{\linewidth}{!}{    
	\begin{tabular}{c|cc|cc|cc}
        \toprule[1pt]
        ~ & \multicolumn{2}{c|}{NYU} & \multicolumn{2}{c|}{ICVL} & \multicolumn{2}{c}{Hands2019-Task1} \\
        \midrule
        Number of views & UNIFORM & RANDOM & UNIFORM & RANDOM& UNIFORM & RANDOM \\
        \midrule
        1 view & 7.93  & 8.81 & 5.56 & 5.96 & 14.39 & 16.18 \\
        3 views & 7.14 & 7.33 & 5.27 & 5.28 & 13.67 & 13.97 \\
        9 views & 6.77 & 6.68 & 4.96 & 4.93 & 12.81 & 13.07 \\
        15 views & 6.55 & 6.52 & 4.85 & 4.88 & 12.61 & 12.92 \\
        25 views & 6.40 & 6.41 & 4.79 & 4.83 & 12.55 & 12.83 \\
        \bottomrule[1pt]
	\end{tabular}
	}
	\caption{Comparison of mean joint error using uniform sampling and random sampling on NYU, ICVL and Hands2019-Task1. "UNIFORM" denotes using uniformly sampled views. "RANDOM" denotes using randomly sampled views.}
	\label{error_uniform_random} 
\end{table*}

\section{2. More Experiments}
\subsection{2.1 Comparison between Uniform Sampled Views and Random Sampled Views}
In our paper, we use uniform sampling on a sphere to obtain the candidate virtual views. 
In order to investigate the effect of random sampled views on hand pose estimation, we compare the hand pose estimation performance with randomly sampled virtual views and uniformly sampled views on a sphere. 
For the experiments of randomly sampled views, the views are randomly sampled from the zenith angle in $[-\pi/3,\pi/3]$ and the azimuth angle in $[-\pi/3,\pi/3]$ on the sphere, and we adopt the same number of virtual cameras as uniformly sampled views.

Table~\ref{error_uniform_random} compares the mean joint error of uniform sampling and random sampling on NYU, ICVL and Hands2019-Task1. We can observe that the performance with random sampled views is inferior to the performance with uniform sampled views.
Conceptually, the patterns of random sampled views are more complex than those of uniform sampling, and make the confidence prediction of random sampled views very hard. Therefore, we choose to use uniform sampling to obtain the candidate virtual views. 

\subsection{2.2 Detailed Results on the ICVL Dataset }
\subsubsection{More Comparisons with State-of-the-art Methods}
\begin{figure*}
\centering 
\includegraphics[width=\linewidth]{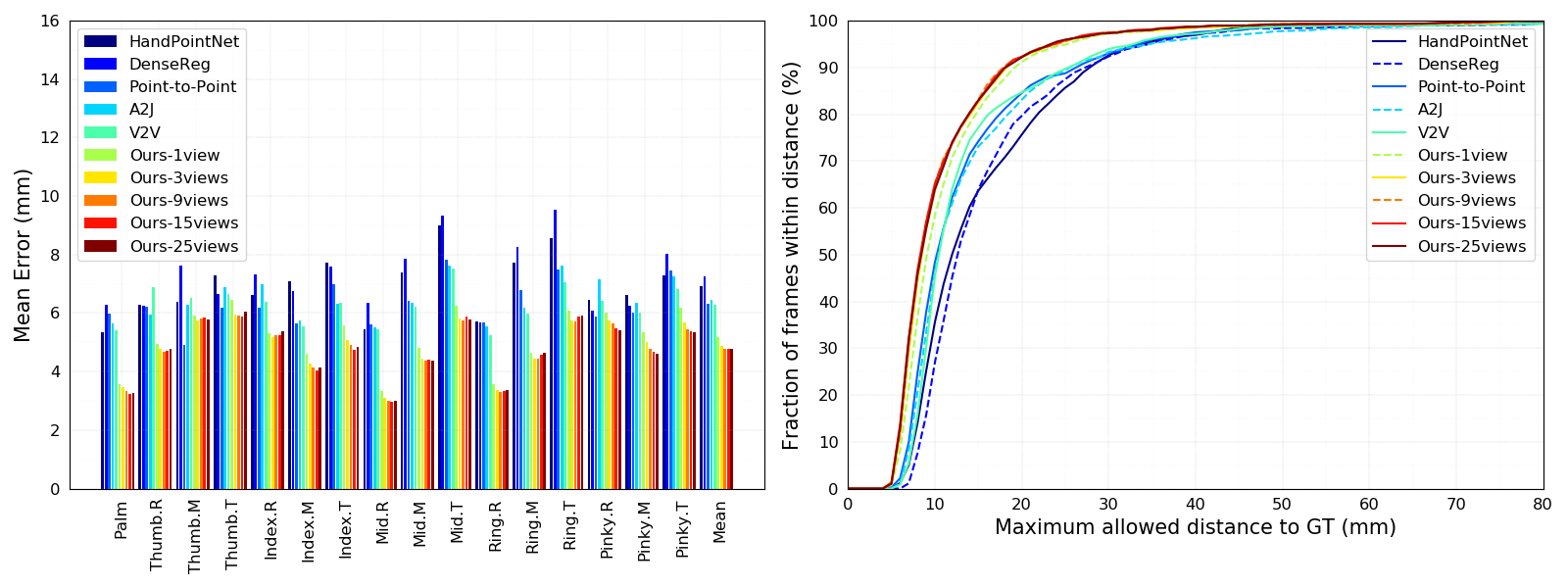}
\caption{Comparison of our proposed method with state-of-the-art methods on the ICVL dataset. Left: 3D distance errors per hand joints. Right: the percentage of success frames over different error thresholds.}
\label{fig:icvl_error}
\end{figure*}

Fig.~\ref{fig:icvl_error} shows the error of each joint and the percentage of success frames over different error thresholds. Since the best prediction results by AWR on ICVL dataset are not released, we do not show AWR \cite{huang2020awr} in Fig.~\ref{fig:icvl_error}. We can observe that our proposed method outperforms the other methods by a large margin on the ICVL dataset.

\begin{figure*}[ht]
    \centering
    \resizebox{\linewidth}{!}{  
    \begin{tabular}{ccccccccc}
      \includegraphics[width=0.09\linewidth]{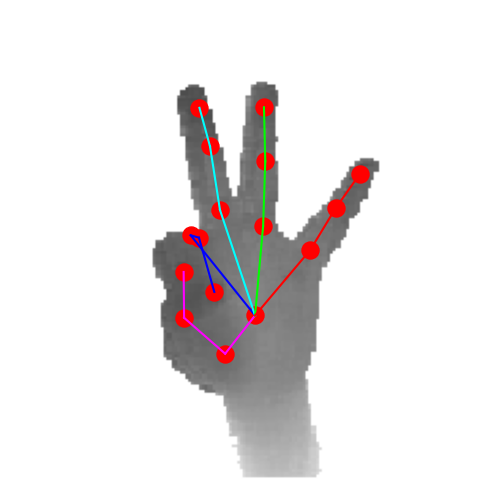}   &  \includegraphics[width=0.09\linewidth]{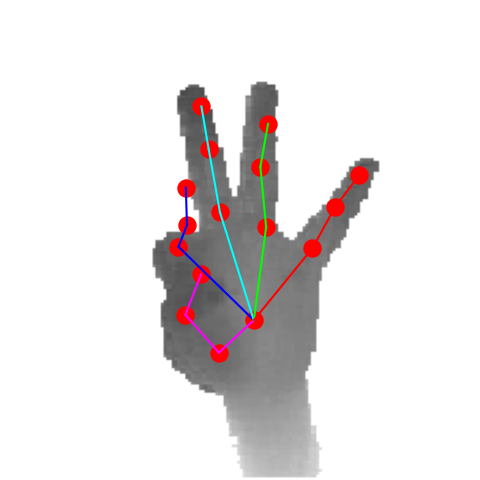} & \includegraphics[width=0.09\linewidth]{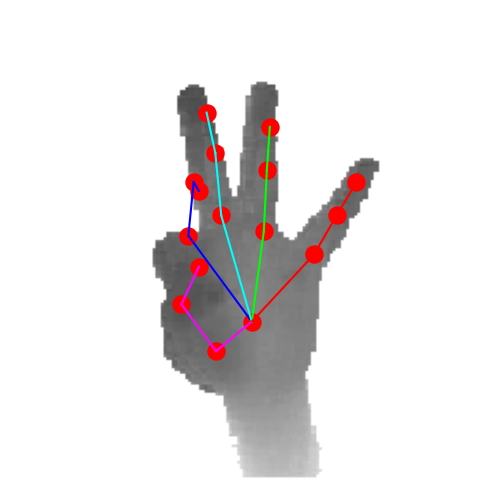} & 
      \includegraphics[width=0.09\linewidth]{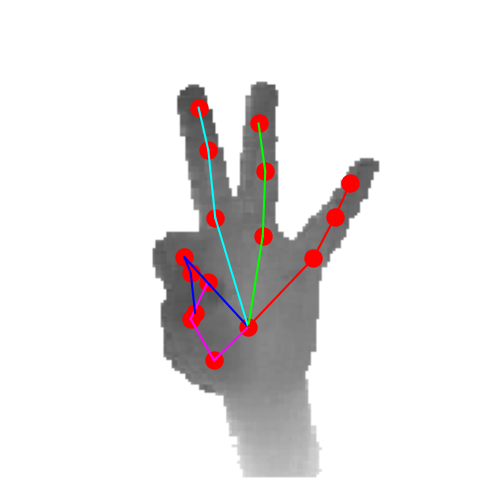}& 
      \includegraphics[width=0.09\linewidth]{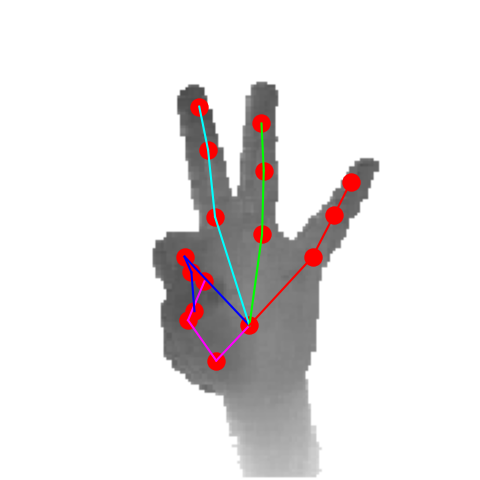}& 
      \includegraphics[width=0.09\linewidth]{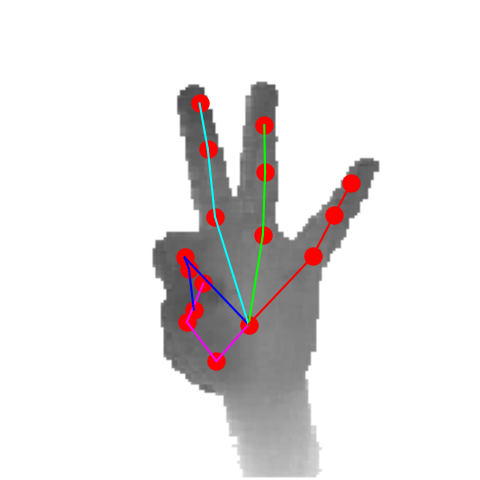}& 
      \includegraphics[width=0.09\linewidth]{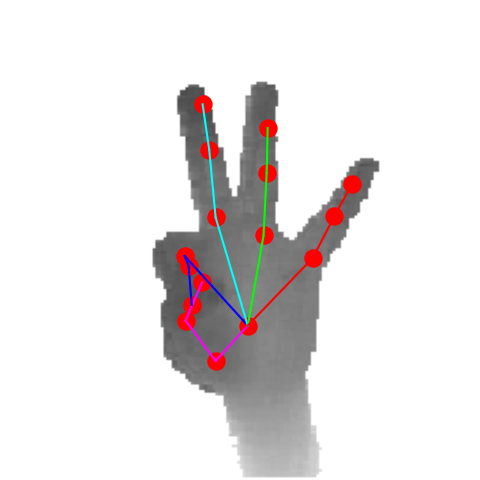}& 
      \includegraphics[width=0.09\linewidth]{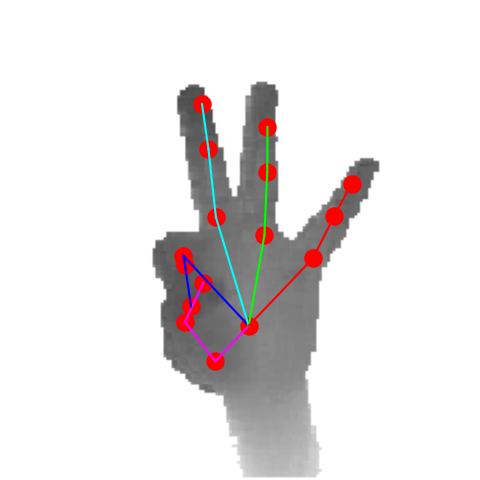}&
      \includegraphics[width=0.09\linewidth]{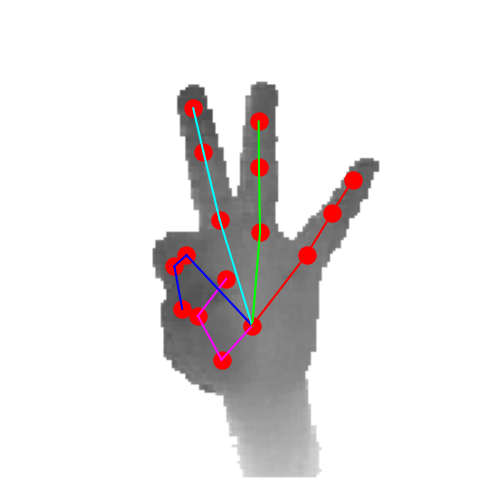} \\
      \includegraphics[width=0.09\linewidth]{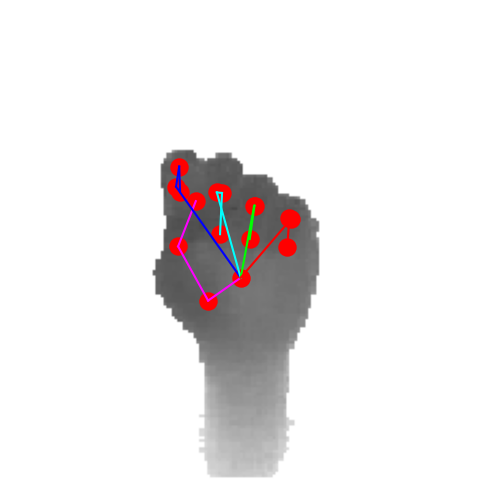}   & \includegraphics[width=0.09\linewidth]{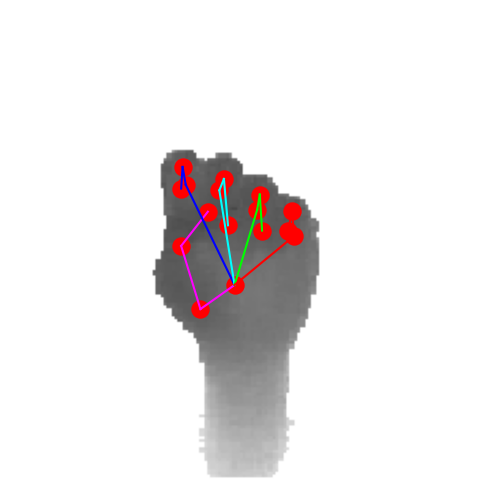}   & \includegraphics[width=0.09\linewidth]{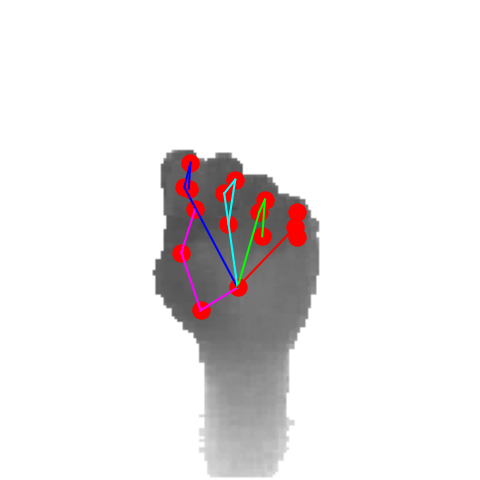}  &  \includegraphics[width=0.09\linewidth]{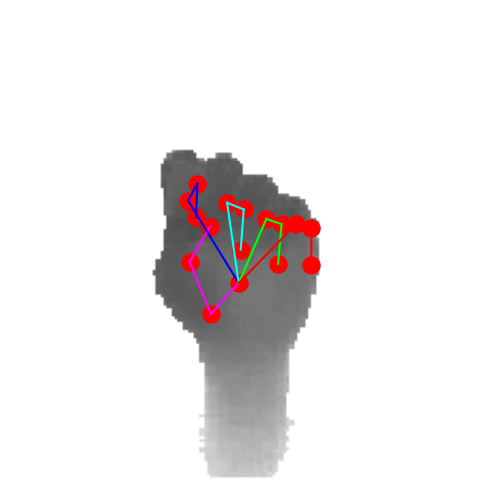}& 
      \includegraphics[width=0.09\linewidth]{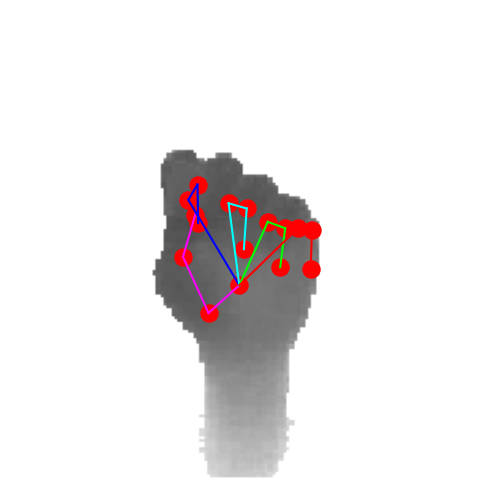}& 
      \includegraphics[width=0.09\linewidth]{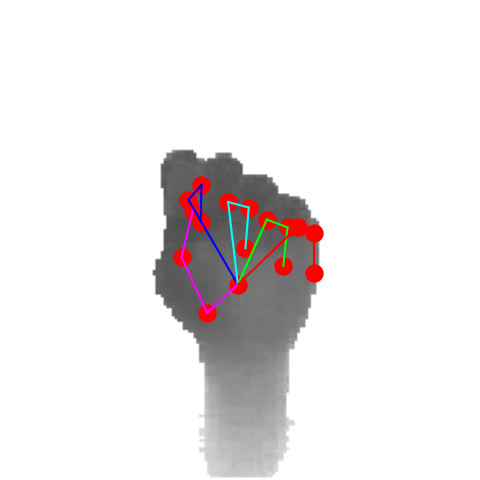}& 
      \includegraphics[width=0.09\linewidth]{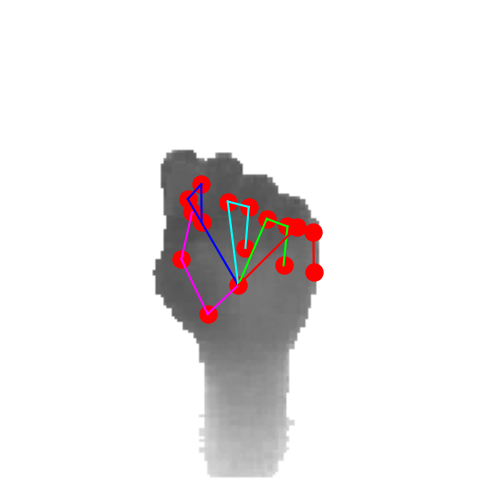}& 
      \includegraphics[width=0.09\linewidth]{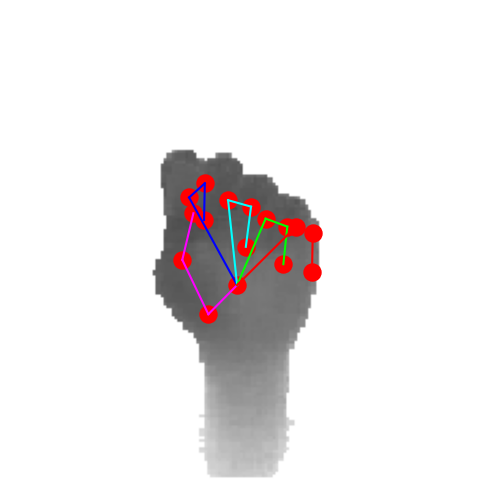}&
      \includegraphics[width=0.09\linewidth]{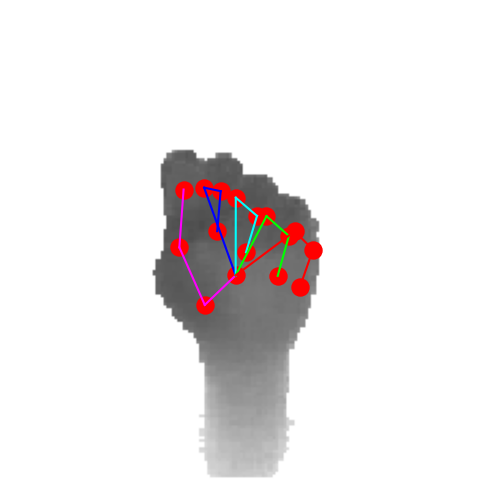} \\ 
      \includegraphics[width=0.09\linewidth]{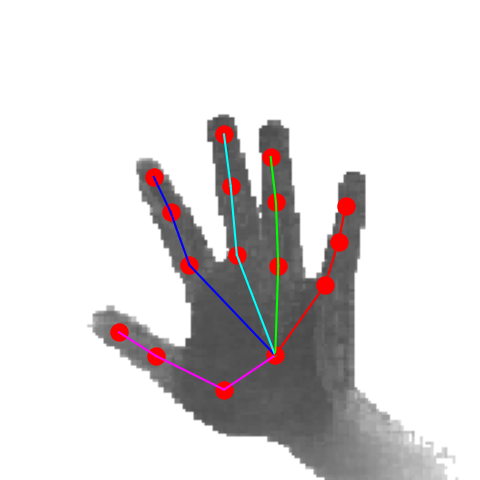} &\includegraphics[width=0.09\linewidth]{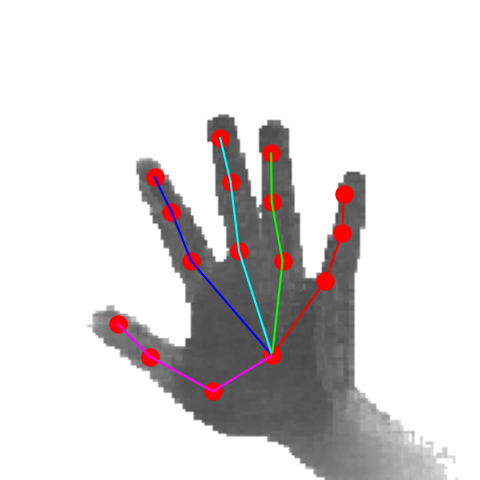} 
      &\includegraphics[width=0.09\linewidth]{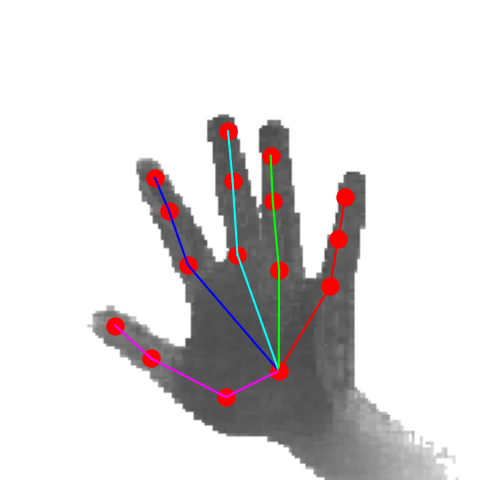} & \includegraphics[width=0.09\linewidth]{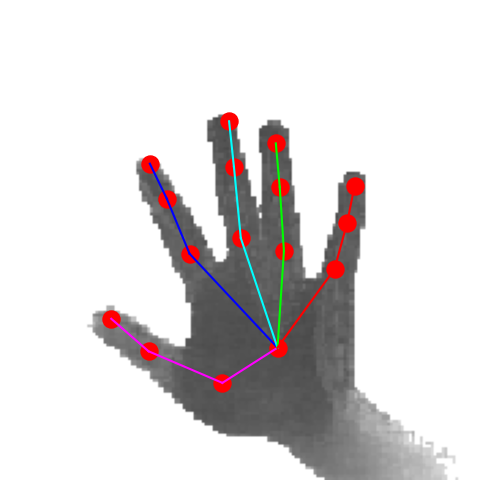}& 
      \includegraphics[width=0.09\linewidth]{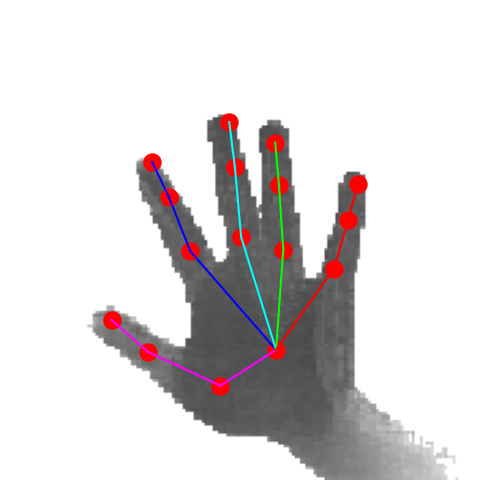}& 
      \includegraphics[width=0.09\linewidth]{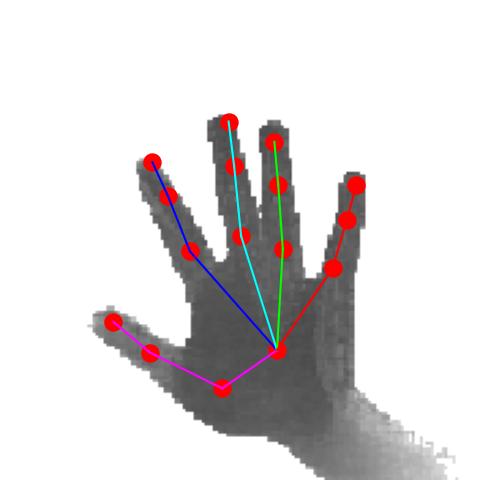}& 
      \includegraphics[width=0.09\linewidth]{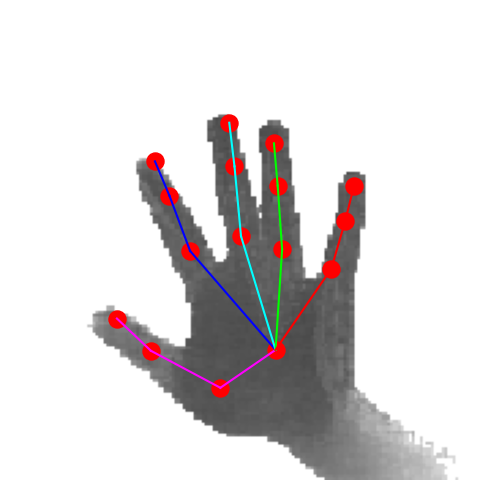}& 
      \includegraphics[width=0.09\linewidth]{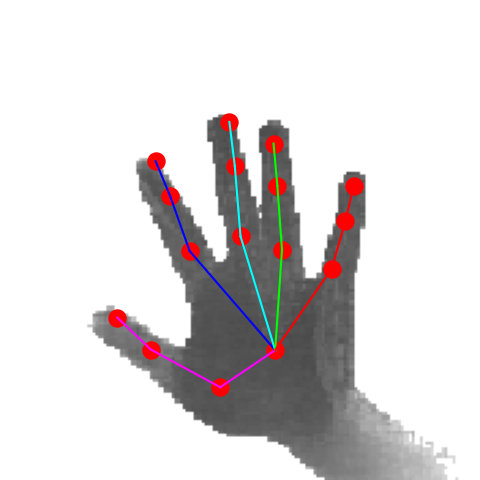}&
      \includegraphics[width=0.09\linewidth]{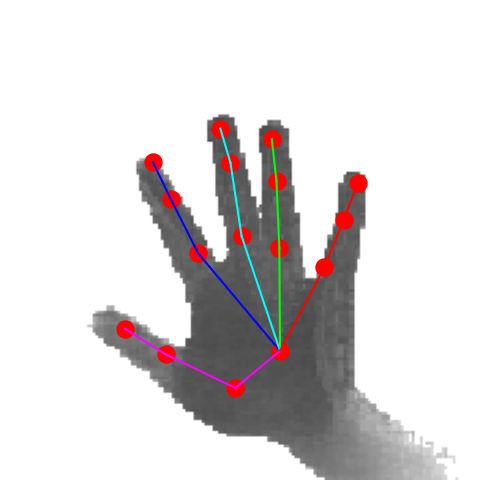} \\ 
      P2P & V2V & A2J & 1 view & 3 views & 9 views& 15 views & 25 views & Ground truth
    \end{tabular}
    }
    \caption{Comparison visualization results with state-of-art methods on ICVL dataset. 
    ``1 view'', ``3 views'', ``9 views'' and ``15 views''  are the results of our method with selected 1, 3, 9 and 15 views from 25 uniformly sampled views, respectively.
    ``25 views'' denotes the results of our method with 25 uniformly sampled views.
    }
    \label{fig:comp_STOA}
\end{figure*}

Fig.~\ref{fig:comp_STOA} shows the qualitative comparison to the SoTA methods on the ICVL dataset.
Especially, our method performs better at the fingertips than the other methods, e.g. the 1st row, and also performs well under severe occlusions, e.g. the 2nd row.

\subsubsection{Effect of View Number}
\begin{figure}[ht]
\centering 
\includegraphics[width=\linewidth]{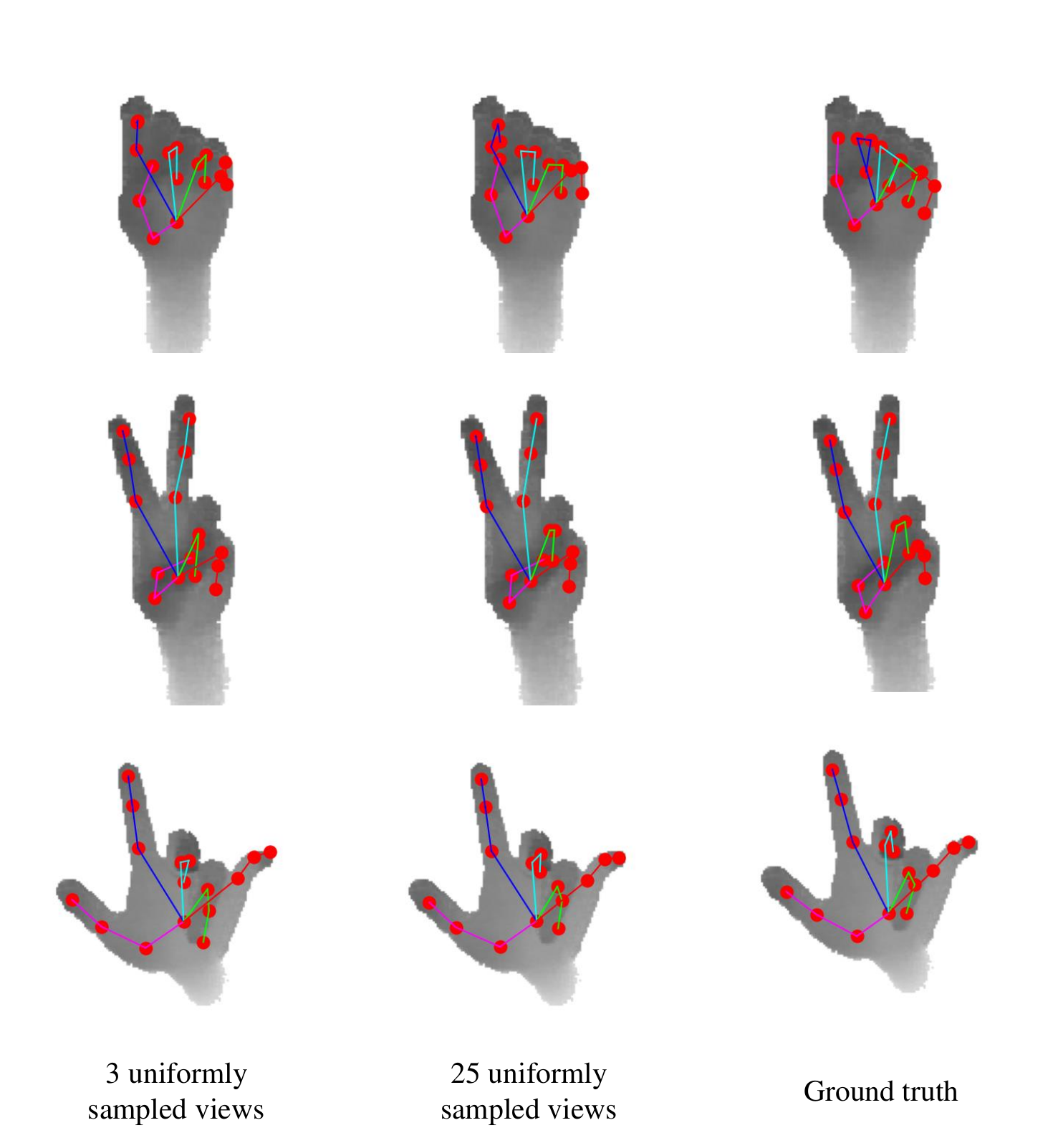}
\caption{Qualitative comparison of our method using different numbers of views on the ICVL dataset.}
\vspace{45em}
\label{fig:comp_views}
\end{figure}

Fig.~\ref{fig:comp_views} shows the results of our method using different numbers of views on the ICVL dataset.
We can observe that using 25 uniformly sampled views can achieve better hand pose estimation performance than using 3 uniformly sampled views. Especially, the estimated fingertip joints using 25 uniformly sampled views are better. 

